\documentclass[journal]{IEEEtran}

\usepackage{ragged2e}

\usepackage[table,xcdraw]{xcolor}
\usepackage{graphicx}
\usepackage{array}
\usepackage{amssymb}
\usepackage{amsthm}
\usepackage{algorithm}
\usepackage{algpseudocode}
\usepackage{amsmath}
\usepackage{multirow}

\ifCLASSINFOpdf
\else
\fi

\begin{document}

%
\title{Deep Reinforcement Learning based Dynamic Optimization of Bus Timetable}


\author{Guanqun~Ai,
        Xingquan~Zuo,~\IEEEmembership{Senior~member,~IEEE},
        Gang~chen,
        and Binglin Wu
\thanks{Xingquan Zuo, Guanqun Ai and Binglin Wu are with School of Computer Science, Beijing University of Posts and Telecommunications, Beijing 100876, China, and also with the Key Laboratory of Trustworthy Distributed Computing and Service, Ministry of Education, Beijing 100876, China.}
\thanks{Gang Chen is with School of Engineering and Computer Science, Victoria University of Wellington, Wellington 6140, New Zealand.}
}

\markboth{\ }%
{Shell \MakeLowercase{\textit{et al.}}:Deep Reinforcement Learning based Dynamic Optimization of Bus Timetable}
%




\IEEEtitleabstractindextext{%
\begin{abstract}
Bus timetable optimization is a key issue to reduce operational cost of bus companies and improve the service quality. Existing methods use exact or heuristic algorithms to optimize the timetable in an offline manner. In practice, the passenger flow may change significantly over time. Timetables determined in offline cannot adjust the departure interval to satisfy the changed passenger flow. Aiming at improving the online performance of bus timetable, we propose a Deep Reinforcement Learning based bus Timetable dynamic Optimization method (DRL-TO). In this method, the timetable optimization is considered as a sequential decision problem. A Deep Q-Network (DQN) is employed as the decision model to determine whether to dispatch a bus service  during each minute of the service period. Therefore, the departure intervals of bus services are determined in real time in accordance with passenger demand. We identify several new and useful state features for the DQN, including the load factor, carrying capacity utilization rate, and the number of stranding passengers. Taking into account both the interests of the bus company and passengers, a reward function is designed, which includes the indicators of full load rate, empty load rate, passengers' waiting time, and the number of stranding passengers. Building on an existing method for calculating the carrying capacity, we develop a new technique to enhance the matching degree at each bus station.  Experiments demonstrate that compared with the timetable generated by the state-of-the-art bus timetable optimization approach based on a memetic algorithm (BTOA-MA), Genetic Algorithm (GA) and the manual method, DRL-TO can dynamically determine the departure intervals based on the real-time passenger flow, saving 8$\%$ of vehicles and reducing 17$\%$ of passengers' waiting time on average.
\end{abstract}

}

\maketitle
\IEEEdisplaynontitleabstractindextext



%

\section{Introduction}
%
%
%
%
\IEEEPARstart{D}{riven}  by fast world-wide economical development, the number of private cars has increased sharply, which leads to the problem of traffic congestion, exhaust emissions and air pollution [2]. The public transportation system has the advantages of large carrying capacity, convenience, efficiency and energy conservation, which can alleviate these problems effectively. However, the bus transportation system in many cities has poor service quality with long waiting time, low punctuality, and high crowdedness during peak hours, which results passengers preferring to drive private cars. Therefore, improving the service quality and operating efficiency of the public transport system is an effective way to attract passengers to ride buses.

Bus timetable optimization is vital to improve the service quality and operating efficiency, in order to public transportation a favorable choice for a large proportion of passengers. It takes into account the interests of both passengers and bus companies, and sets the frequency of bus departures according to passenger flow to reduce the operational cost of the bus company while meeting the demand of passengers [8]. Current bus timetable optimization methods mainly include heuristic algorithms [1-10], [15-18] and mathematical programming algorithms [11-14]. Those methods typically divide a day into several time periods, and then optimize the departure interval based on the historical passenger flow in a purely offline fashion. Once the departure intervals (timetable) are determined, the bus system will strictly follow the fixed timetable, which cannot be adjusted easily according to the real-time passenger flow. However, in practice, sudden changes of passenger flow occur frequently [19]. In this case, the timetable cannot satisfy the actual need of passengers.

Reinforcement learning is one of the major intelligent decision-making methods [20-22]. It imitates humans to learn from its experiences obtained while interacting with an unknown environment, with the goal to identify the optimal strategy through the continuous trial, error and evaluation.
In this paper, we propose a deep reinforcement learning (DRL) based bus timetable dynamic optimization method (DRL-TO) with the aim to optimize departure interval (frequency) adaptively and in real time. Different from existing methods for bus timetable optimization, we take every minute during the bus operation period as a decision point, and apply DRL to decide whether to dispatch new bus service at every decision point. By this way, the departure interval (frequency) can be determined adaptively, enabling progressive and dynamic generation of the timetable. Thus, we transform the bus timetable optimization problem into a real-time decision-making problem. A DRL framework based on rule-constrained DQN is designed. We select the load factor, carrying capacity utilization efficiency, the number of stranding passengers as the state features of the reinforcement learning framework, and ``departure" and ``no departure" as the optional actions. Taking into account the interests of the bus company and passengers, a reward function is designed to capture comprehensively a variety of decision factors, including load rate, empty load rate, waiting time and the number of stranding passengers. To the best of our knowledge, no published work has ever exploited reinforcement learning to dynamically determine the departure interval of bus services (departure frequency) in real time. Our paper is the first work to introduce reinforcement learning to dynamically optimize bus timetable.
The contributions of this paper are as follows:

1)	We formulate the first time in literature the bus timetable optimization is a real-time decision-making problem, and devise a new DRL based method to tackle this problem.

2)	We propose several new and useful state features and a new reward function that can provide informative feedback to the DRL system.

3)	We propose a new optimization objective and develop a new method to calculate the carrying capacity in association with the new objective.

4)	Experiments demonstrate that DRL-TO outperforms the state-of-the-art bus timetable optimization methods, and has a high potential for practical use with good performance.

The remainder of the paper is organized as follows. In section 2, we give a brief introduction to the optimization methods of bus timetable and the application of reinforcement learning in the field of transportation. Section 3 details the bus timetable optimization problem. Section 4 presents the proposed approach. And the experimental results for three real datasets are presented in Section 5. Finally, Section 6 draws conclusions.

\section{Related works}

This section focuses on the existing bus timetable optimization methods and the application of reinforcement learning in the field of traffic timetable optimization.

\subsection{bus timetable optimization}
The bus timetable defines the planned departure time for a series of bus to leave the starting station during a full service period (e.g., from 5am to 10pm on a single day) [18]. For some existing algorithms, such as [1-14], the departure time should always be kept uniform. Alleviating this restriction, other research works, such as [15-18] can provide non-uniform departure interval.

The uniform interval timetable is mainly generated by mathematical programming algorithms and heuristic algorithms where the latter algorithms are more widely used [1-8]. There are many research works using genetic algorithm (GA) or its improved versions to optimize the departure timetable. Han \emph{et al}. [1] leveraged GA to optimize the objective function composed of the sum of the costs paid by passengers on the bus, the sum of the costs paid by passengers to transfer and the variable operator payment. Zhou \emph{et al}.  [2] established a scheduling model based on the optimization objectives of passenger waiting time and bus system operating costs, and improved GA with tabu search algorithm to solve this problem. A bus timetable optimization model for improving the quality of bus service has been proposed by Lu \emph{et al}. [3]. The authors employed the Non dominated Sorting Genetic Algorithm -II (NSGA-II) to search for Pareto-optimal solutions based on this model. Wihartik \emph{et al}. [4] presented an improved GA and established an integer programming model to solve the bus timetable optimization problem. A hybrid algorithm combining GA and Simulated Annealing (SA) is further proposed by Tang \emph{et al}. [5]. Li \emph{et al}. [6] established a dual-objective optimization model to minimize the total travel time and the total waiting time of all passengers. A new GA based algorithm with variable length chromosomes is designed to solve the problem. Gkiotsalitis \emph{et al}.  [7] considered the uncertainty of travel time, passenger actual need, and applied an improved GA to generate robust offline timetables. A dual-objective optimization model is established by Yang \emph{et al}.  [8] to minimize the total waiting time of passengers and the costs of bus company, and applied the Global Positioning System (GPS) trajectory data of bus and passenger information to optimize the key parameters or variables in the optimization model. In addition, an improved NSGA-II was designed to quickly search for Pareto solutions.

In addition to GA, there are also other heuristic algorithms for bus timetable optimization. An integer programming model and two heuristic methods are proposed by Oudheusden \emph{et al}.  [9] to conduct offline optimization of the bus timetables in Bangkok. Ceder \emph{et al}.  [10] applied graphical heuristics to examine different strategies regarding their efficacies in creating near-optimal bus timetables.

Mathematical programming has also been exploited for solving bus timetable optimization problems with uniform departure intervals. Ceder \emph{ et al}.  [11] proposed four methods for bus departure timetable optimization based on passenger survey data. Two of them are based on point test (maximum load) data and the other two are based on travel check (load curve) data. A bus departure interval control model was designed by Sun \emph{et al}.  [12] considering both passengers' travel costs and bus companies' operating costs. A transition model of bus departure interval was presented by Dong \emph{et al}.  [13] that integrates passenger demand and traffic congestion, which is applied to optimize the bus departure interval under traffic congestion. Aiming at improving passenger satisfaction and bus operation efficiency, Shang \emph{et al}. [14] employed several numerical search methods to optimize bus timetable.
The non-uniform interval departure timetable can support varying departure intervals in each time period. Usually this category of methods does not simultaneously optimize the bus timetable of multiple time periods, but focuses on one period at a time. If we consider one day as a time period, the parallel genetic algorithm [15] and the bus timetable optimization approach based on a memetic algorithm (BTOA-MA) [18] can be utilized to optimize the entire bus timetable for a whole day. Sun \emph{et al}.  [16] and Li \emph{et al}.  [17] instead considered one or two hours as a time period.

In summary, the current research of bus timetable optimization methods, including uniform and non-uniform intervals, focuses primarily on offline optimization. When the actual passenger flow changes, the departure timetable generated by the above methods cannot adapt quickly to match the changed passenger flow.

\subsection{Reinforcement learning technology in transportation systems}

The application of reinforcement learning in transportation field is mainly divided into two aspects: (1) Reinforcement learning in bus service; (2) Optimization of train and subway timetables.


\subsubsection{Reinforcement learning in bus service}

The recent literature provides two interesting applications of DRL to bus service management: (1) to optimize the relevant parameters of other algorithms [25][26]; and (2) to optimize parking time after the timetable is determined[27]. Hassane \emph{et al}.  [25] presented a mathematical model to optimize the bus timetable. They introduce a meta-heuristic method to solve the model, in which reinforcement learning is exploited to adaptively select genetic operators of the method. Matos \emph{et al}.  [26] formulated the bus timetable optimization problem as boolean satisfiability problem (SAT), and reinforcement learning is used to optimize SAT parameters. Additionally, a multi-deep reinforcement learning framework was proposed by Wang \emph{et al}.  [27] to optimize the parking time at each bus station. The departure timetable is not optimized by this method.
In this paper we develop new reinforcement learning methods to techniques to dynamically optimize bus timetables. As far as we know, similar research works have never been reported in the literature.

\subsubsection{Optimization of train and subway timetables}

To the best of our knowledge, only Zou \emph{et al}. [28] proposed a reinforcement learning method to optimize the subway departure interval in 2006. This method employed SARSA as the intelligent algorithm, which exploited the position of all vehicles, the number of passengers and the running direction of the light rail as the state space. By making a series of ``departure" and ``no departure" decisions, the algorithm was shown to produce an optimal departure timetable that is capable of optimizing jointly several objectives, including the waiting time of passengers, travel time, passenger satisfaction, etc. The subway timetable optimization is completely different from bus timetable optimization in the following aspects: (1) The subway runs on a fixed track and is less affected by traffic conditions. We can make ideal expectations on the running time between any two stations. However, the traffic condition must be considered explicitly in bus timetable optimization, making it a more complicated problem. (2) The number of passengers taking the subway is relatively large. Compared with the carrying capacity utilization efficiency of the vehicle, we are more concerned about the quality of service (waiting time of passengers). However, we must consider the interests of both the passengers and the bus company. In summary, the state space and reward function of the bus timetable optimization problem are more complex than the subway timetable optimization problem. The method in [28] cannot be applied directly to bus timetable optimization.

Reinforcement learning is also applied to the problem of the train stop time optimization [29] and Train Timetable Rescheduling (TTR) [30-33].  The train timetabling aims at determining, for each line involved, the arrival and departure times of all operating trains at each station [29]. During train management, a timetable may deviate from the planned one under uncertain disturbances which can be caused by the failure of equipment and system along the railway line. Notably, the TTR problem concerns about how to re-acquire a feasible timetable by rerouting, reordering, retiming and canceling trains within a short period of time [30-33]. These problems are substantially different from the problem of bus timetable optimization.

\section{The bus timetable optimization problem}

The optimization of the bus timetable aims to consider the interests of both passengers and the bus company, and set the departure time of buses to meet the demand of passenger flow.

The interests of passengers are mainly reflected in the service level of buses. The main quantitative indicators are bus congestion and the waiting time of passengers, while the interests of bus companies are mainly affected by the number of departures (departure intervals) in the timetable. When the bus departure interval is too large, the carrying capacity provided by the timetable cannot meet the passengers' demand, resulting in overcrowding, poor riding experience, large number of stranding passengers, and poor service quality. On the contrary, when the departure interval is too small, the vehicle will be empty, and the operating costs will increase. Therefore, adequately determining the departure frequency according to the needs of passengers and bus companies is of great significance for saving bus operating costs and improving the quality of service.

\section{Dynamic optimization of bus timetable based on deep reinforcement learning}

The bus line generally has two directions: up and down. In each direction, there is a departure control point (the starting station), which controls the departure of buses in that direction. As shown in Fig.1, one direction of a bus line is taken as an example to describe the dynamic process of the bus timetable optimization. The square box represents the departure station, the five-pointed star represents the last station, and the circle represents the intermediate station. The bottom horizontal axis is the timeline. Every minute represents a decision point, and the green one is the departure time.

To adjust the departure interval in real time, we need to design a controller which decides whether to departure or not according to the real-time need of passengers and traffic condition at every decision point. We take every minute during the bus operation period as a decision point. In this way, we transform the bus timetable optimization problem into a sequential decision problem.

\begin{figure}[htbp]
\centering
\includegraphics[height=8cm,width=8.5cm]{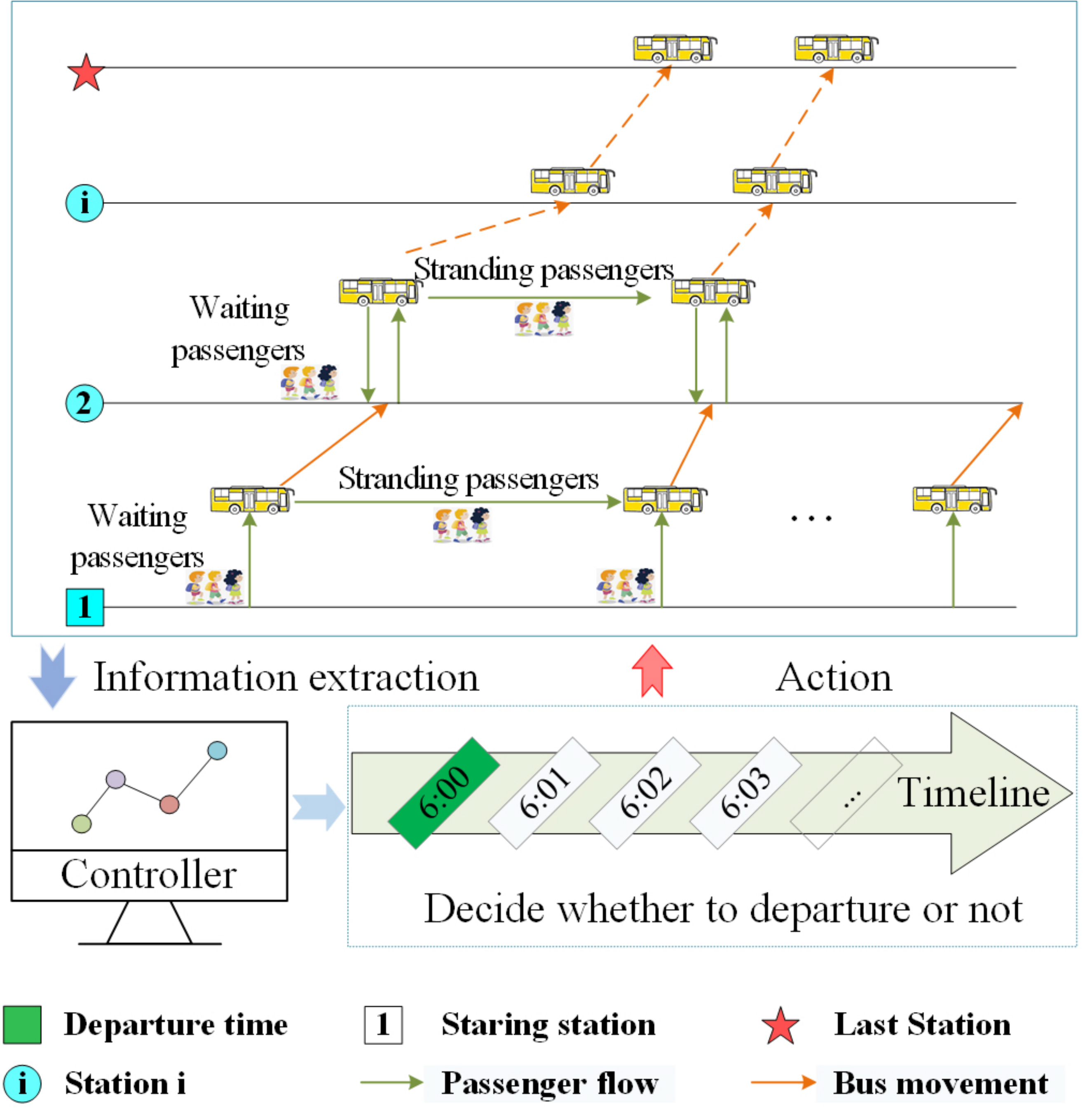}
\centering
\caption{The bus timetable optimization problem.}
\end{figure}

The need of passengers can be reflected in the passenger flow between each pair of stations in the bus line, which can be represented by an Origin-destination matrix (OD matrix). $O D \in \mathbb{R}^{m_{o} \times k_{o} \times m_{d} \times k_{d}}$ represents the boarding time, ${m_d}$ represents the get off time, ${k_o}$ represents the get on station, ${k_d}$ represents the get off station. The traffic conditions along the bus line can be captured by the transit time between two adjacent stations. In particular, $t_m^k$ represents the time required to travel from the ${k-th}$ station to the ${(k+1)-th}$ station for the bus depart in the ${m-th}$  minute.

The sequential decision problem can be modeled as a discrete-time Markov Decision Process   (MDP), which lay the theoretical foundation for DQN. The corresponding MDP model comprises of 5 components that jointly form a tuple $< S,A,P,R,\gamma  >$. In particular, $S$  refers to the entire state space where every state $s$   in $S$. We use $A$   to denote the state at time step $t$  and ${s_{t+1}}$  to denote the new state obtained by taking action  $a$ in state $s_t$. $A$ represents action space including two actions: ``departure" and ``no departure". We present ``departure" as 1, and ``no departure" as 0.  $r$ denotes the rewards, that provide immediate performance feedback to the DRL system in consideration of full load rate, empty load rate, waiting time, and the number of stranding passengers. All these MDP components can significantly affect the reinforcement learning performance. Their detailed designs are introduced in the following subsections.

\subsection{Framework of deep reinforcement learning for timetable optimization}

The common intellige nt decision-making methods for MDP include supervised learning methods, decision-making methods based on expert knowledge, and reinforcement learning methods [23]. Different from the first two types of approaches that rely either on a large amount of labeled training data or expert knowledge that is difficult to extract.

As a primary method for intelligent decision making, deep reinforcement learning can automatically learn a policy without using any manually prepared training data or human expertise [24]. In this paper, DRL is exploited to adaptively determine the next bus service departure time, enabling bus timetables to be generated dynamically according to passenger flows.

\begin{figure}[htbp]
\centering
\includegraphics[height=4cm,width=8.5cm]{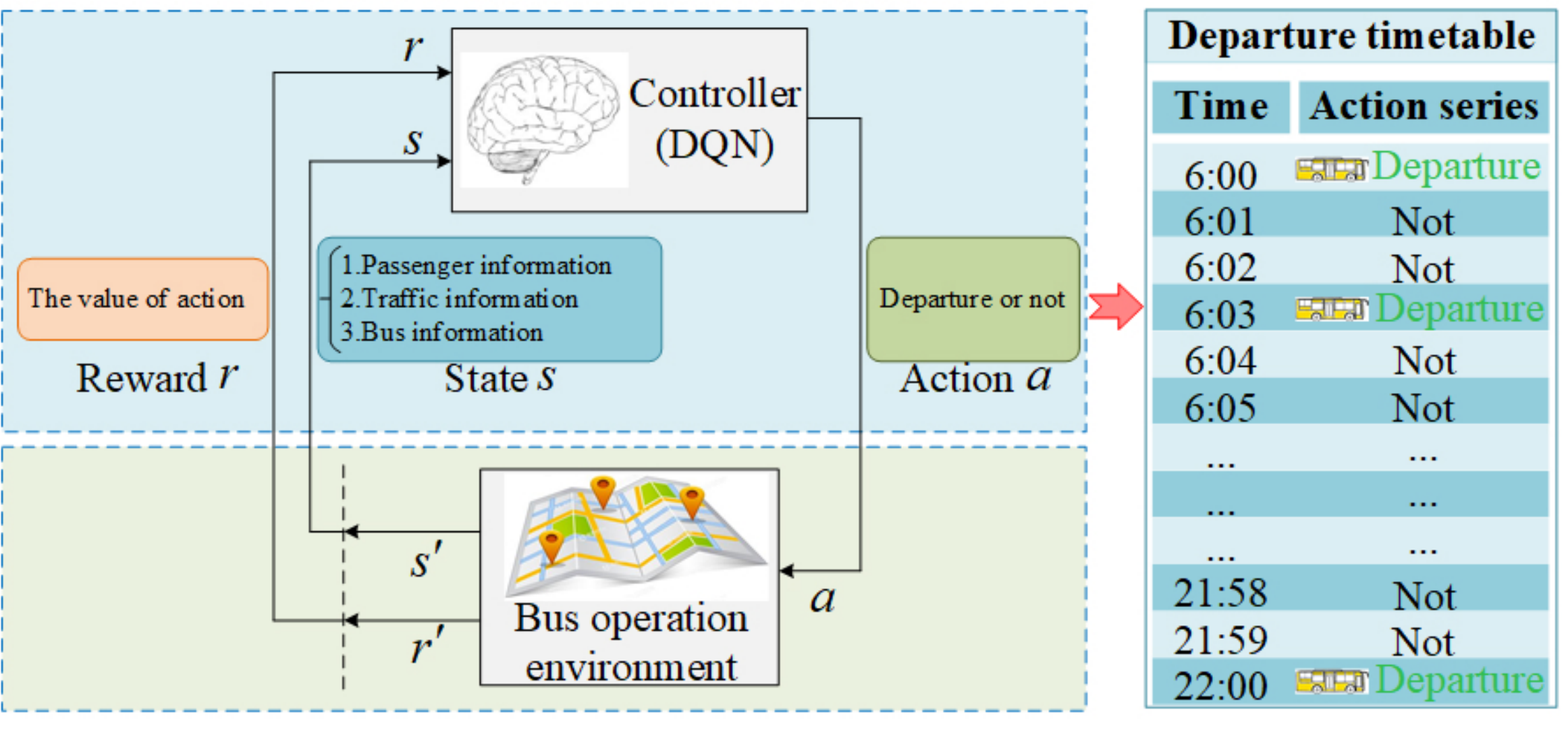}
\centering
\caption{Framework of the proposed method in this paper.}
\end{figure}

The framework of bus timetable optimization based on DRL is illustrated in Fig.2. The decision-making process of reinforcement learning is as follows: Starting from the first minute of bus service period (the   departure time of the first and last bus does are pre-determined by the bus company), the controller (agent) decides the actions to take (``departure" or ``no departure") after every passed minute. In association with each performed action, a state change in the bus system will be witnessed (passenger information, traffic information, bus information, etc.). Meanwhile an immediate reward is provided to the controller (DRL agent). The controller (agent) subsequently decide a new action in the next minute based on the new state. The system repeats the above process until the end time of the bus timetable.

As the number of bus line stations increases, more state related factors must be considered to make suitable bus departure decisions. This will increase the state space, resulting in the problem of dimensionality disaster. DQN [34] is a widely explored and highly successful DRL method. It combines deep neural networks with Q-Learning and can work effectively on a large state space. Therefore, we choose DQN for departure decision-making.

\subsection{Deep Q-Network}

DQN has two major improvements: (1) Through the continuous interaction with it's environment, DQN obtains a series of environment samples in the form of $(s,a,r,s')$  and saves them in the experience replay buffer. During every training cycle, a portion of the data is randomly selected from the experience replay buffer as the training samples. This method breaks the correlation among environment samples, resulting in reliable learning performance[35]; (2) In order to solve the problem that the single value network overestimates action value, DQN introduces the target network to evaluate the state-action value [36].

We apply DQN to train a neural network based model of the value function, which will guide action selection by the controller (agent). The value function captures the expected long-term benefits of any given state-action pair, and is defined by the Optimal Bellman equation below.

\begin{equation}\label{equ1}
{q^ * }(s,a) = \sum\limits_{s',r} {p(s',r|s,a)} [r + \gamma {\max _{a'}}{q^*}(s',a')]
\end{equation}
where $p(s',r|s,a)$ represents the probability of getting a reward $r$  and transferring to a new state $s'$  after taking action $a$  in state $s$, ${q^* }(s,a)$  indicates the maximum value the current state-action pair to be evaluated. $a'$  represents the action performed in $s'$.

We use the combination of expert knowledge and $\varepsilon-greedy$  strategy[38] to limit the DQN exploration process. We limit the DQN action exploration process according to the maximum interval  ${T_{\max }}$ and minimum interval ${T_{\min }}$, either provided by human experts based on their practical experience or determined according to the demands of the bus company. In this paper, we analyze the bus timetables produced by human experts for real-world bus services determine ${T_{\min }}$  and  ${T_{\max }}$. When the current departure interval is between ${T_{\min }}$  and  ${T_{\max }}$, the  $\varepsilon-greedy$ strategy is adopted. If the current departure interval ${t_{ml}}$  is less than the minimum interval ${T_{\min }}$,  any new bus departure is disallowed; if  ${t_{ml}}$ is greater than the maximum interval ${T_{\max }}$, a bus will immediately departure from its starting station. The pseudo code is as follows:

\begin{algorithm}[htb]
  \caption{ Action selection of heuristic DQN.}
  \begin{algorithmic}[1]
    \Require
     $ \varepsilon$

    \State If random() $< \varepsilon :$
    \State ${\kern 1pt} {\kern 1pt} {\kern 1pt} {\kern 1pt} {\kern 1pt}$  if ${t_{ml}} > {T_{\max }}:$
    \State $ {\kern 1pt} {\kern 1pt} {\kern 1pt}{\kern 1pt} {\kern 1pt} {\kern 1pt} {\kern 1pt} {\kern 1pt} {\kern 1pt} {\kern 1pt} {\kern 1pt} {\kern 1pt} {\kern 1pt} a \leftarrow 1$
    \State ${\kern 1pt} {\kern 1pt} {\kern 1pt} {\kern 1pt} {\kern 1pt}  elseif {t_{ml}} < {T_{\min }}:$
    \State $ {\kern 1pt} {\kern 1pt} {\kern 1pt}{\kern 1pt} {\kern 1pt} {\kern 1pt} {\kern 1pt} {\kern 1pt} {\kern 1pt} {\kern 1pt} {\kern 1pt} {\kern 1pt} {\kern 1pt} a \leftarrow 0$
    \State ${\kern 1pt} {\kern 1pt} {\kern 1pt} {\kern 1pt} {\kern 1pt}  else :$
    \State $ {\kern 1pt} {\kern 1pt} {\kern 1pt}{\kern 1pt} {\kern 1pt} {\kern 1pt} {\kern 1pt} {\kern 1pt} {\kern 1pt} {\kern 1pt} {\kern 1pt} {\kern 1pt} {\kern 1pt} a \leftarrow random(s)$

    \State $ Else :$
    \State ${\kern 1pt} {\kern 1pt} {\kern 1pt} {\kern 1pt} {\kern 1pt}$  if ${t_{ml}} > {T_{\max }}:$
    \State $ {\kern 1pt} {\kern 1pt} {\kern 1pt}{\kern 1pt} {\kern 1pt} {\kern 1pt} {\kern 1pt} {\kern 1pt} {\kern 1pt} {\kern 1pt} {\kern 1pt} {\kern 1pt} {\kern 1pt} a \leftarrow 1$
    \State ${\kern 1pt} {\kern 1pt} {\kern 1pt} {\kern 1pt} {\kern 1pt}  elseif {t_{ml}} < {T_{\min }}:$
    \State $ {\kern 1pt} {\kern 1pt} {\kern 1pt}{\kern 1pt} {\kern 1pt} {\kern 1pt} {\kern 1pt} {\kern 1pt} {\kern 1pt} {\kern 1pt} {\kern 1pt} {\kern 1pt} {\kern 1pt} a \leftarrow 0$
    \State ${\kern 1pt} {\kern 1pt} {\kern 1pt} {\kern 1pt} {\kern 1pt}  else :$
    \State $ {\kern 1pt} {\kern 1pt} {\kern 1pt}{\kern 1pt} {\kern 1pt} {\kern 1pt} {\kern 1pt} {\kern 1pt} {\kern 1pt} {\kern 1pt} {\kern 1pt} {\kern 1pt} {\kern 1pt} a \leftarrow \mathop {\arg \max }\limits_a Q(s,a)$
  \end{algorithmic}
\end{algorithm}

\subsection{Reward function design}

The reward function is essential for our DQN controller (agent) to learn how to optimize bus timetable effectively. In this paper, we propose a new optimization goal which takes into account the interests of both passengers and the bus company.  The interests of bus companies are mainly reflected in the utilization of carrying capacity. The interests of passengers are mainly reflected in the waiting time and the probabilities for them to be stranded. Stranding contributes significantly to poor user experience. This is not only because it will significantly increase the waiting time, but also cause strong anxiety among stranded passengers due to Murphy¡'s law[37].

There are only two actions for bus control points. If the action is always 0, the bus will not departure, which can help the bus company to save costs but cause passengers to wait for a long time. Therefore, we use  $1 - ({o_m}/{e_m})$   as a reward for action 0. ${o_m}$  is the actual need of the carrying capability at the ${m-th}$  minute, ${e_m}$  is the carrying capability of a vehicle departing at the ${m-th}$  minute. ${o_m}/{e_m}$  is the \emph{capacity consumption rate}. The smaller the capacity consumption rate of the bus, the higher the reward for action 0. When the action is 1, it will help improve the service level of the bus. But if it is always 1, it will increase the bus vacancy rate, causing high bus operation costs. Therefore, we use ${o_m}/{e_m}$  as the reward for action 1. The greater the capacity consumption rate of the bus, the greater the reward for action 1. $\omega  \times {W_m}$  is used as a penalty for waiting time.  ${W_m}$ is the sum of waiting time with respect to all passengers who ride on the bus departing at the ${m-th}$  minute. When the action is 1, the waiting time for passengers riding this bus will not increase, so we only penalize action 0. In order to improve the quality of service and avoid the situation of stranding many passengers, we impose a stricter penalty on the number of stranding passengers. When there are passengers stranded at the station, the reward value of the action will drop sharply. In summary, we design the reward function as shown in Eq. (2):

\begin{equation}\label{equ2}
{r_m}({s_m},a) = \left\{ {\begin{array}{*{20}{c}}
{\left[ \begin{array}{l}
1 - ({o_m}/{e_m}) - \\
(\omega  \times {W_m}) - (\beta  \times d{s_m})
\end{array} \right]}&{,\;\;\;a = 0}\\
{({o_m}/{e_m}) - (\beta  \times d{s_m})}&{,\;\;\;a = 1}
\end{array}} \right.
\end{equation}
where the state  ${s_m}$ is defined in Eq.(4),  ${ds_{m}}$  is the number of stranding passengers at the $m-th$ minute, which is defined in Eq.(3). $\omega $   and $\beta$  are used to adjust the penalty weight for waiting time and the number of stranding passengers. Because the passenger flow of each line is different, $\omega $  is different for different lines. The number of stranding passengers  ${ds_{m}}$ is an a non-negative integer. When  ${ds_{m}}$ is greater than zero, strong penalty must be applied to Eq. (2). For this purpose, we let  $\beta = 0.2 $. When  ${ds_{m}}$ is more than 10\% of the maximum passenger capacity of the vehicle, whether the reward is positive or negative depends on  ${ds_{m}}$.

\begin{equation}\label{equ3}
d{s_m} = \sum\limits_{k = 1}^{K - 1} {d{s_{mk}}}
\end{equation}
where  ${ds_{mk}}$ is the number of stranding passengers in time $m$  (i.e., the $m-th$  minute of a single day) due to bus reached its capacity limit at the $k-th$ station.

\subsection{State space design}

It is crucial to select informative features to form the system state for effective DRL. As mentioned above, the reward function involves the actual need of carrying capability  ${o_m}$, the carrying capability of a vehicle departing  ${e_m}$, the sum of waiting time among all passengers  ${W_m}$, and the number of stranding passengers  ${ds_m}$. The design of the above features is based on the assumption of the current departure. That is to say, assuming that the bus departs at the current decision point, the value of each feature is calculated through passenger flow prediction. The passenger flow prediction algorithm proposed in  [38][39] can achieve high prediction accuracy. We hence use this algorithm directly in this paper. Therefore, we consider the above features to represent every state in the state space. Clearly, the corresponding feature values can vary substantially in different time periods. In view of this, we introduce a temporal dimension into the state space. We normalize every state feature into the range between 0 and 1 before passing them as input to the DQN. The final state definition is shown in Eq. (4).

\begin{equation}\label{equ4}
{s_m} = [{x_{1m}},{x_{2m}},{x_{3m}},{x_{4m}},{x_{5m}},{x_{6m}}]
\end{equation}
where ${x_{1m}}$  and   ${x_{2m}}$  represent normalized time. The respective definitions are shown in Eq. (5) and Eq. (6).

\begin{equation}\label{equ5}
{x_{1m}} = {t_h}\;/\;24
\end{equation}

\begin{equation}\label{equ6}
{x_{2m}} = {t_m}\;/\;60
\end{equation}
where ${t_h}$ and ${t_m}$ represent the hour and minutes after the hour, which is another way of expressing the $m-th$  minute of the day.
In order to determine whether there are passengers stranded at any station, we employ the maximum full load rate as state, and its definition is shown in Eq. (7).

\begin{equation}\label{equ7}
{x_{3m}} = C_{\max }^m/{C_{\max }}
\end{equation}
where ${C_{\max}}$  represents the maximum number of passengers that each bus can carry, and $C_{\max }^m$  represents the maximum number of passengers on board during the trip of the vehicle that departures at the $m-th$  minute. If ${x_{3m}}$  is 1, it means that the number of bus carried has reached its limit, a clear indication that some passengers may be stranded.

  ${x_{4m}}$  represents the normalized waiting time of all passengers who take the bus departured at the  $m-th$ minute, and its definition is given in Eq. (8):

\begin{equation}\label{equ8}
{x_{4m}} = W_m/{\mu}
\end{equation}
where $W_m$  represents the total waiting time of all passengers who take the bus departured in the $m-th$  minute, as defined in Eq. (9):

\begin{equation}\label{equ9}
{W_m}{\rm{ = }}\sum\nolimits_{k = 1}^{k = (K - 1)} {\sum\nolimits_{i = 1}^{i = {l_{mk}}} {(t_{_b}^{_{_{m,i,k}}} - t_a^{_{_{m,i,k}}})} }
\end{equation}
where  $t_a^{m,i,k}$ is the time that passenger $i$  arrives at the $k-th$  station, and  $t_b^{m,i,k}$ is the time of passenger $i$  gets on the bus at the $k-th$  station.  $\mu$  is the normalized coefficient of waiting time. We have calculated the waiting time of all passengers from the historical data, which is consistently less than 5000 minutes. Based on that, we decide to set   $\mu=5000$. Besides, the respective number of passengers who got on and got off the bus departing in the $m-th$  minute at the  $k-th$  station is denoted respectively by $l_m^k$  and $h_m^k$.

  ${x_{5m}}$ is the carrying capacity consumption rate of the bus departing at the $m-th$  minute, defined below:

\begin{equation}\label{equ10}
{x_{{\rm{5}}m}} = {o_m}/{e_m}
\end{equation}
where ${o_m}$  is the actual need of the carrying capability at the ${m-th}$  minute, ${e_m}$  is the carrying capability of a vehicle departing at the ${m-th}$  minute. We will show detailed calculation of the carrying capacity in the next subsection.

\subsection{Carrying capacity calculation}

Carrying capacity is an index to measure both passengers' actual need and the carrying ability of bus. We proposed a method to calculate carrying capacity in an earlier work  [18]. In this paper, we improve the carrying capacity calculation method in order to express the environmental state features better.

In our previous work [18], carrying capacity is defined as the product of the number of passengers and the distance they traveled. The goal is to narrow the gap between the capacity provided by the departure timetable and the passengers' actual need. In this paper, our goal is to reduce the gap without stranding passengers. To avoid passenger stranded, we require to dispatch it as long as the current bus will be departured is predicted to be fully loaded, regardless of the distance between the stations. Therefore, our reward function pays more attention to the number of passengers at each station, and does not consider the distance between stations. We propose a new carrying capacity calculation method based on the number of stations. The carrying capacity provided by a bus,  ${e_m}$, is computed by

\begin{equation}\label{equ11}
{e_m}{\kern 1pt} {\kern 1pt} {\kern 1pt} {\rm{ = }}{\kern 1pt} {\kern 1pt} {\kern 1pt} {\kern 1pt} \alpha  \times C \times {\rm{(}}K - 1)
\end{equation}
where $K$  is the number of stations on the bus line (see Fig.1 for an example);  $C$ is the number of seats in a bus; $\alpha$  is a coefficient reflecting the comfort level of passengers in the bus. We set $\alpha$ to 1.5 based on several existing research [18]. We can calculate the carrying capability provided by the timetable during the time period  $[{t_a},{t_b}]$ by Eq. (12).

\begin{equation}\label{equ12}
\sum\limits_{\{ {t_a} \le m < {t_b}\} } {{e_m}}
\end{equation}

If there is a bus to departure,  ${o_m}$ represents the carrying capacity actual used, which is  defined as the sum of the product of the number of passengers and the number of their riding stations. It is computed by:

\begin{equation}\label{equ13}
\begin{array}{l}
{o_m}{\rm{ = }}\sum\nolimits_{k = 1}^{k = K} {\begin{array}{*{20}{c}}
{(l_m^k - h_m^k){\rm{Number of passengers}} \times }\\
{1{\kern 1pt} {\kern 1pt} {\kern 1pt} {\rm{Number of stations}}}
\end{array}} \\
\;\;\; = \sum\nolimits_{k = 1}^{k = K} {\begin{array}{*{20}{c}}
{(l_m^k - h_m^k)({\rm{Number of passengers}} \times }\\
{{\rm{Number of stations)}}}
\end{array}}
\end{array}
\end{equation}
where $K$   is the number of stations on the line. Similarly, we can calculate the carrying capability consumed during the time period  $[{t_a},{t_b}]$  by Eq. (14).

\begin{equation}\label{equ14}
\sum\limits_{\{ {t_a} \le m < {t_b}\} } {{o_m}}
\end{equation}

\section{Experimental results}

DRL-TO is applied to 3 real-world bus timetables in a Chinese city. Its results are compared against the state-of-the-art bus timetable optimization approach based on a memetic algorithm (BTOA-MA)[18], Genetic Algorithm (GA) and the manual method.

\subsection{Experimental data}

In the experiments, we use real-world swiping data obtained from three live bus lines of 2, 230 and 239 in Xiamen City on June 15, 2018. Each line contains two directions: up and down. There is one control point in each direction. The Information regarding those bus lines can be found in Table \uppercase\expandafter{\romannumeral1}.

In addition, parameters of proposed method (DRL-TO) are presented as follows. The DQN network architecture is based on the work in [40]. Table \uppercase\expandafter{\romannumeral2} shows the parameters used in the model.

\begin{table}[htp]
\caption{Information of bus lines.}
\begin{tabular}{lllll}
\hline
Line ID              & Direction & Service time & Stations & Passengers \\ \hline
\multirow{2}{*}{2}   & Up        & 6:20-22:00   & 37       & 4968       \\
                     & Down      & 6:30-22:00   & 36       & 4515       \\
\multirow{2}{*}{230} & Up        & 6:00-22:00   & 33       & 7739       \\
                     & Down      & 6:45-22:00   & 33       & 6818       \\
\multirow{2}{*}{239} & Up        & 6:00-23.05   & 35       & 5896       \\
                     & Down      & 6:40-23:40   & 36       & 5802       \\ \hline
\end{tabular}
\end{table}

\begin{table}[htp]
\caption{Parameters of DQN.}
\begin{tabular}{ll}
\hline
Parameters              & Values                \\ \hline
Number of hidden layers & 10                    \\
Number of hidden units  & 300                   \\
Initial weight value    & Normal Initialization \\
Activation function     & ReLU                  \\
Learning rate           & 0.01                  \\
Discount rate           & 0.4                   \\
Experience memory size  & 3000                  \\
Batch size              & 32                    \\ \hline
\end{tabular}
\end{table}

\subsection{Experimental result and analysis}

In order to verify the effectiveness of DRL-TO, we conduct experiments in both static and dynamic environment. Besides, we perform ablation study to investigate the validity of the state space and reward function we designed in subsection \uppercase\expandafter{\romannumeral4} \emph{C} and \emph{D}.

Bus timetable obtained by DRL-TO is compared against the solutions generated by GA, BTOA-MA [18] and manually designed bus timetables (Manual) of the three bus lines. BTOA-MA is proposed in [18], GA refers to a variation of BTOA-MA without using the local search technique.  GA and BTOA-MA are coded in C++ language. DRL-TO is coded in pythorch. Experiments are conducted on a PC with 3.20 GHz CPU and 8G RAM. 30 independent runs of BTOA-MA and GA are performed.

\subsubsection{Experimental results in a static environment}

\begin{table}[ht]
\caption{Experimental results of DRL-TO, BTOA-MA, GA, and the Manual timetable.}
\begin{tabular}{llllllll}
\hline
\multicolumn{2}{l}{Lines ID}                                                                             & \multicolumn{2}{l}{2} & \multicolumn{2}{l}{230} & \multicolumn{2}{l}{239} \\ \hline
\multicolumn{2}{l}{Direction}                                                                            & Up        & Down      & Up         & Down       & Up         & Down       \\ \hline
\multirow{3}{*}{Manual}                                                                            & ND  & 72        & 72        & 104        & 110        & 119        & 109        \\
                                                                                                   & AWT & 6.2       & 7         & 5.2        & 7.6        & 4          & 10.6       \\
                                                                                                   & NSP & 0         & 0         & 24         & 611        & 0          & 1635       \\ \hline
\multirow{3}{*}{GA}                                                                                & ND  & 78        & 72        & 99         & 101        & 117        & 110        \\
                                                                                                   & AWT & 5.6       & 6.7       & 4.8        & 6.4        & 4          & 8.4        \\
                                                                                                   & NSP & 0         & 0         & 14         & 370        & 0          & 1048       \\ \hline
\multirow{3}{*}{BTOA-MA}                                                                           & ND  & 66        & 66        & 86         & 92         & 107        & 103        \\
                                                                                                   & AWT & 7.3       & 7.4       & 6.2        & 7.3        & 4.6        & 9.7        \\
                                                                                                   & NSP & 0         & 20        & 50         & 574        & 0          & 1207       \\ \hline
\multirow{3}{*}{\textbf{\begin{tabular}[c]{@{}l@{}}DRL-TO(D)\end{tabular}}} & ND  & 65        & 65        & 80         & 90         & 101        & 94         \\
                                                                                                   & AWT & 6.3       & 6.8       & 5.6        & 6          & 4.3        & 6          \\
                                                                                                   & NSP & 0         & 0         & 0          & 21         & 0          & 77         \\ \hline
\multirow{3}{*}{\textbf{\begin{tabular}[c]{@{}l@{}}DRL-TO(W)\end{tabular}}}      & ND  & 71        & 70        & 88         & 99         & 111        & 107        \\
                                                                                                   & AWT & 5.8       & 6.1       & 4.7        & 5.1        & 3.8        & 5.2        \\
                                                                                                   & NSP & 0         & 0         & 0          & 3          & 0          & 77         \\ \hline
\end{tabular}
\end{table}

Table \uppercase\expandafter{\romannumeral3} summarizes the experimental results obtained by GA, BTOA-MA, Manual and DRL-TO for both the upward and downward directions of the three bus lines. The number of departures (ND), average waiting time (AWT)  and the number of stranding passengers (NSP) are compared. AWT is measured in minutes. We report two experimental results with respect to priority optimization of waiting time (DRL-TO(W)) and priority optimization of departure times (DRL-TO(D)). The respective results are enclosed in brackets in Table \uppercase\expandafter{\romannumeral3}. Obviously, DRL-TO(D) outperforms Manual and GA in terms of all the three performance indicators. DRL-TO(W) outperforms BTOA-MA in terms of all the three performance indicators. The departure timetable generated by DRL-TO produces clearly less number of bus departures, the shortest passenger waiting time and minimum number of stranding passengers. Therefore, our DQN method has the best overall performance.


\begin{figure}[htbp]
\centering
\includegraphics[height=6.5cm,width=8cm]{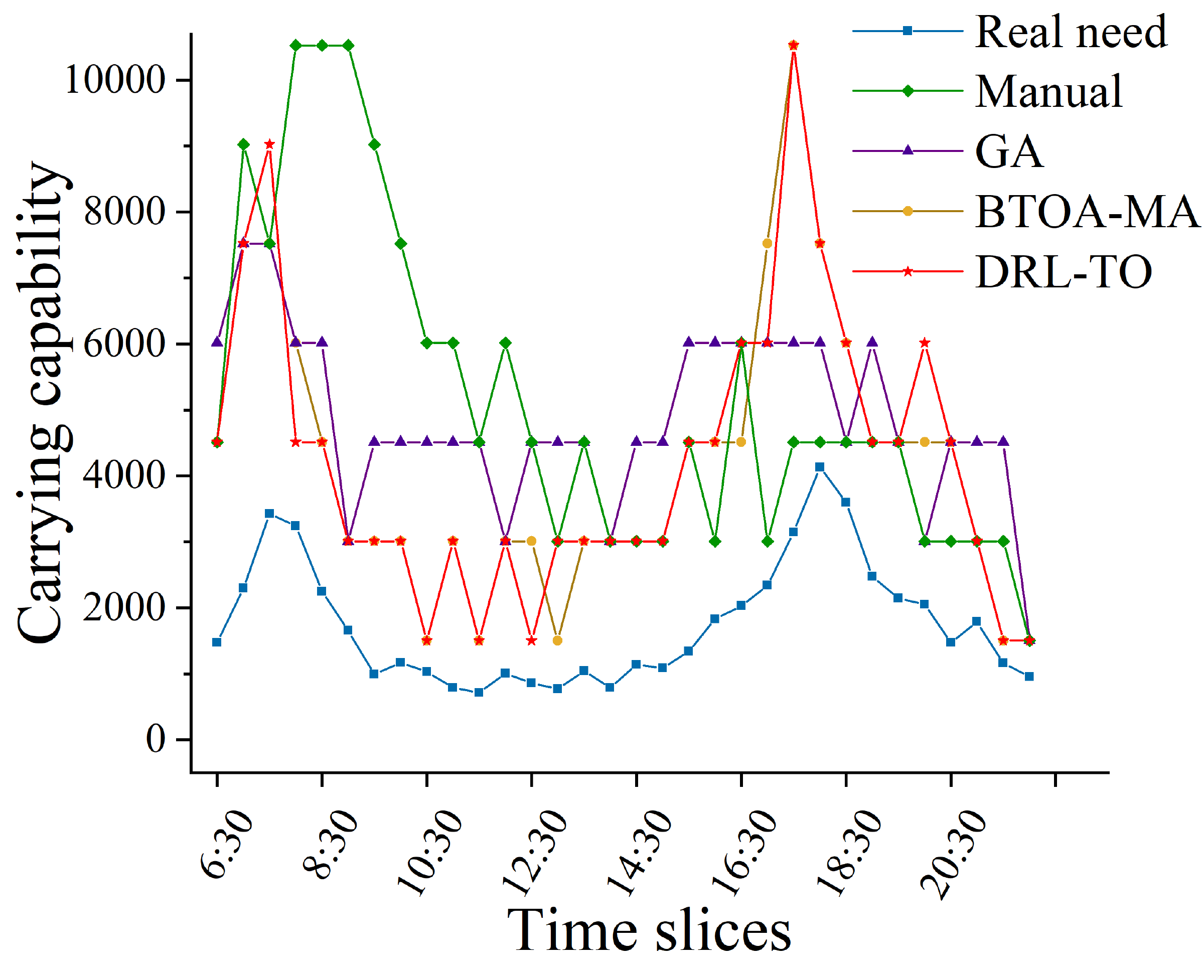}
\centering
\caption{Actual need of passengers on carrying capability vs. carrying capability provided by the optimized timetable of downward direction of bus line 230.}
\end{figure}

\begin{figure}[htbp]
\centering
\includegraphics[height=6.7cm,width=8cm]{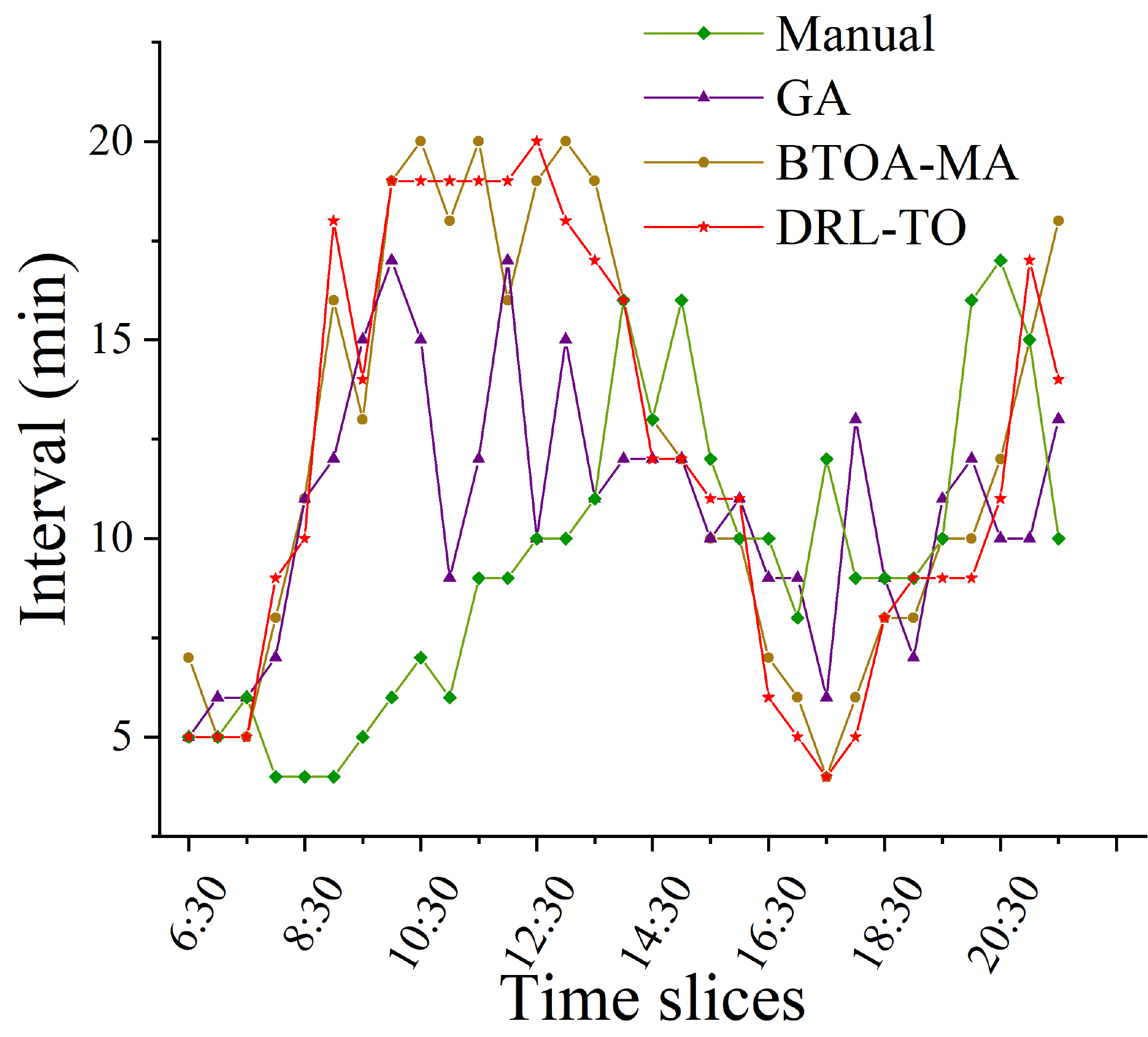}
\centering
\caption{The average interval of the departure timetable of downward direction of bus line 230 generated by DRL-TO, GA, BTOA-MA and Manual in each half hour.}
\end{figure}

Fig. 3 compares the carrying capability of the timetable for the downward direction of bus line 230 obtained by DRL-TO, GA, BTOA-MA, Manual and the actual required carrying capability. The carrying capacity at each time point represents the capacity required or provided by the timetable in the next half an hour, which can be calculated by the Eq. (12) and Eq. (14). As shown in Fig.3, DRL-TO uses less capacity to cover the actual need capacity, which can also be confirmed from Fig.4. Compared with other methods, DRL-TO is more sensitive to changes in passenger flow, and can effectively adjust departing interval according to actual needs. When the passenger flow is small (large), the departure interval is large (small). Although BTOA-MA uses the least capacity, in other words, the smallest number of departures, it cannot adequately cover all actual needs. This is clearly evidenced by the longest average waiting time and the largest number of stranding passengers in Table \uppercase\expandafter{\romannumeral3}.

\subsubsection{Experimental results in a dynamic environment}

In order to verify the ability of DRL-TO to cope with sudden changes of passenger flows, we simulate the change of passenger flow in two aspects: (1) advance or delay the time at which passenger flow peak arrives; (2) Random sample the passenger's card swiping data to reduce or increase the peak passenger flow. As shown in Fig. 5, the two peaks of the actual passenger flow have been shifted in tune. For the morning peak, we delay it by 0.5, 1.5, and 2.5 hours respectively, and for the afternoon peak, we advance them by 1, 2 and 3 hours, and delay 1 and 2 hours respectively.

\begin{figure}[htbp]
\centering
\includegraphics[height=6cm,width=8cm]{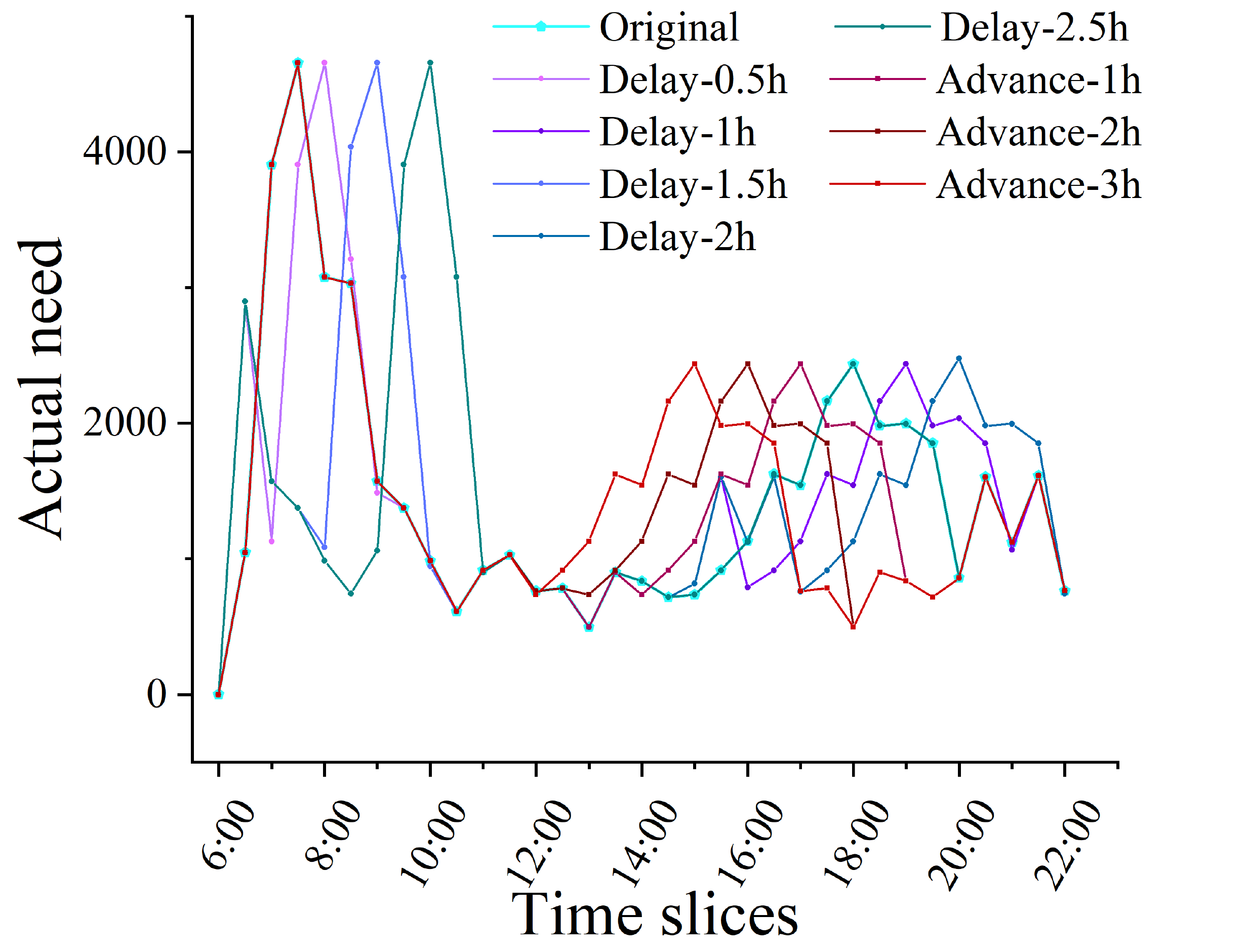}
\centering
\caption{Different time shifts on the peak of passenger flow for the upward direction of bus line 230 in time dimension.}
\end{figure}

We study the ability of DRL-TO, GA, BTOA-MA and Manual to cope with passenger flow changes in time. As shown in Fig. 6 (a) and (b), we shift the peak of 15:00-18:00 in the afternoon of line 230 upward direction for three hours, and move the passenger flow from 12:00-15:00 to 17:00-20:00. The departure timetable generated by GA, BTOA-MA and Manual can't quickly adapt to this change of the passenger flow. It can be seen from Fig. 6 (d) that there is a notable mismatch between the peak value of the carrying capacity provided by those methods and the actual passenger need. From the comparison between Fig. 6 (f) and Fig.6 (e), it is obvious that when the carrying capacity demands for immediate change, DRL-TO can respond quickly and effectively. This ensures that any temporal changes to the peak have no noticeable impact on  service quality achievable by DRL-TO. When the peak of carrying capacity is moving, the times of departures the waiting time of passengers and number of stranding passengers of DRL-TO is not changed significantly. This can also be verified from Table \uppercase\expandafter{\romannumeral4}: when the passenger flow changes, the average waiting time for passengers and the number of stranding passengers of other algorithms increase. The greater the peak shifts in time, the longer the waiting time for passengers and more stranding passengers. Experiment results in Table \uppercase\expandafter{\romannumeral4} also confirm that DRL-TO can make adjustments to maintain service quality when the peak of passenger flow shifts in time.

\begin{table*}[ht]
\caption{Experimental results of DRL-TO, BTOA-MA, GA, and the Manual timetable to the peak moving of actual need carrying capacity of upward direction of bus line 230.}
\begin{tabular}{
>{\columncolor[HTML]{FFFFFF}}l
>{\columncolor[HTML]{FFFFFF}}l
>{\columncolor[HTML]{FFFFFF}}l
>{\columncolor[HTML]{FFFFFF}}l
>{\columncolor[HTML]{FFFFFF}}l
>{\columncolor[HTML]{FFFFFF}}l
>{\columncolor[HTML]{FFFFFF}}l
>{\columncolor[HTML]{FFFFFF}}l
>{\columncolor[HTML]{FFFFFF}}l
>{\columncolor[HTML]{FFFFFF}}l }
\hline
\multicolumn{2}{l}{\cellcolor[HTML]{FFFFFF}}            & Delay 0.5h & Delay  1.5h & Delay  2.5h & Delay  1h & Delay  2h & Advance  1h & Advance  2h & Advance  3h \\ \hline
\cellcolor[HTML]{FFFFFF}                          & ND  & 104        & 104         & 104         & 104       & 104       & 104         & 104         & 104         \\
\cellcolor[HTML]{FFFFFF}                          & AWT & 5.2        & 6           & 5.8         & 5.3       & 5.1       & 5.2         & 5.2         & 5.2         \\
\multirow{-3}{*}{\cellcolor[HTML]{FFFFFF}Manual}  & NSP & 7          & 199         & 142         & 24        & 24        & 24          & 24          & 24          \\ \hline
\cellcolor[HTML]{FFFFFF}                          & ND  & 99         & 99          & 99          & 99        & 99        & 99          & 99          & 99          \\
\cellcolor[HTML]{FFFFFF}                          & AWT & 5.1        & 5.2         & 5.3         & 4.9       & 5         & 4.8         & 4.8         & 4.9         \\
\multirow{-3}{*}{\cellcolor[HTML]{FFFFFF}GA}      & NSP & 18         & 21          & 41          & 14        & 2         & 14          & 14          & 14          \\ \hline
\cellcolor[HTML]{FFFFFF}                          & ND  & 86         & 86          & 86          & 86        & 86        & 86          & 86          & 86          \\
\cellcolor[HTML]{FFFFFF}                          & AWT & 7.6        & 8.8         & 9.1         & 7.2       & 7.4       & 7.29        & 7.32        & 7.6         \\
\multirow{-3}{*}{\cellcolor[HTML]{FFFFFF}BTOA-MA} & NSP & 149        & 510         & 565         & 50        & 39        & 50          & 50          & 50          \\ \hline
\cellcolor[HTML]{FFFFFF}                          & ND  & 86         & 86          & 84          & 81        & 86        & 94          & 87          & 90          \\
\cellcolor[HTML]{FFFFFF}                          & AWT & 5.1        & 5           & 5.1         & 5.6       & 5         & 4.5         & 5.1         & 4.7         \\
\multirow{-3}{*}{\cellcolor[HTML]{FFFFFF}DRL-TO}  & NSP & 0          & 0           & 0           & 0         & 0         & 0           & 0           & 0           \\ \hline
\end{tabular}
\end{table*}

In order to get the real-time sudden change of the passenger flow size, we randomly sample the card swiping data for the upward direction passengers of line 2. The sampling rates are 0.5, 0.7, 0.9, 1.1, 1.3, 1.5, and 1.7 respectively. The sampling result is shown in Fig.7.
\begin{figure}[h]
\centering
\includegraphics[height=12cm,width=9cm]{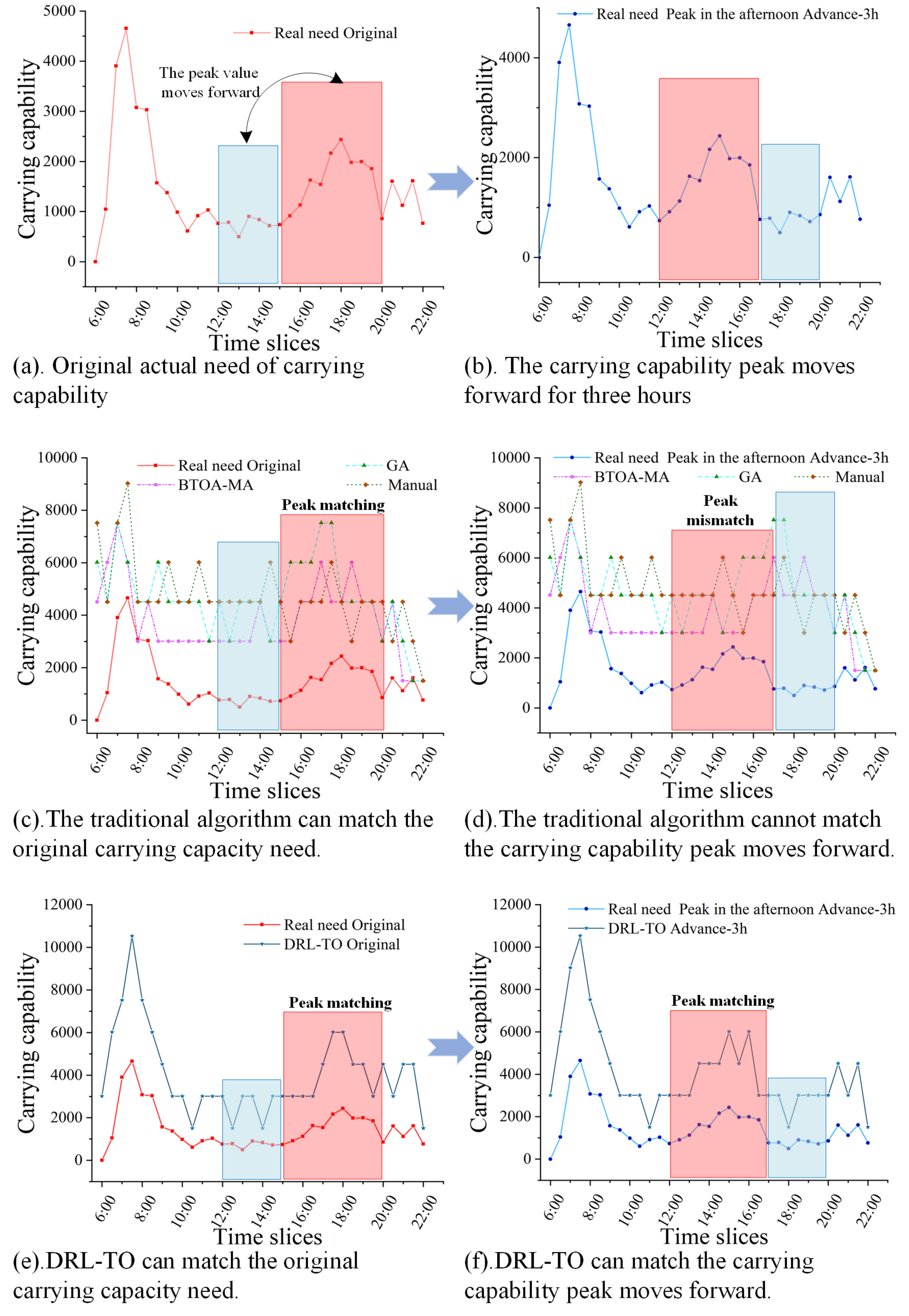}
\centering
\caption{Different time shifts on the peak of passenger flow for the upward direction of bus line 230 in time dimension.}
\end{figure}

\begin{figure}[htbp]
\centering
\includegraphics[height=6cm,width=8cm]{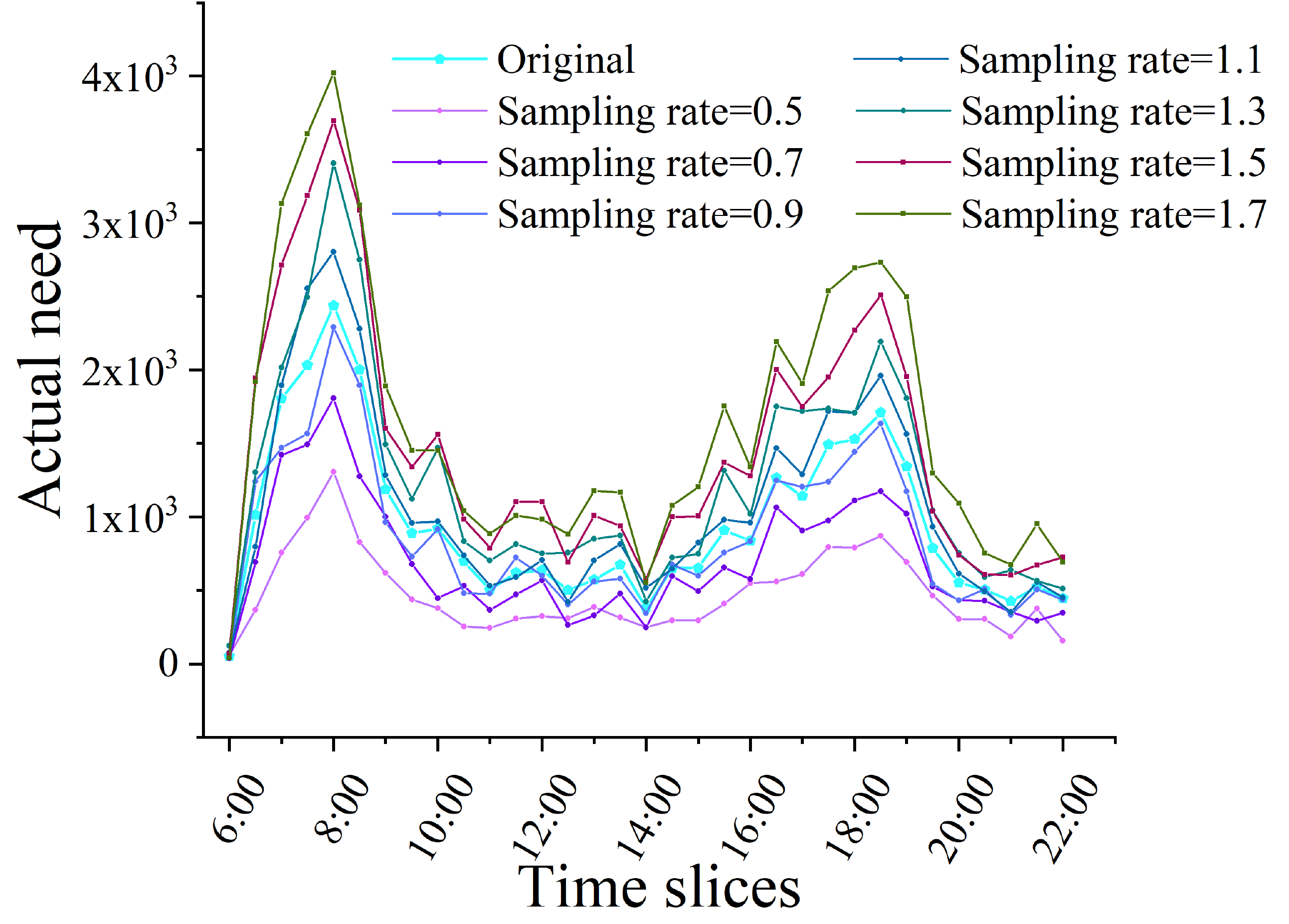}
\centering
\caption{The influence of parameter $\omega$ on the number of departures and waiting time.}
\end{figure}

\begin{table}[ht]
\caption{Experimental results of DRL-TO, BTOA-MA, GA, and the Manual timetable for a range of sampling rates regarding the actual need carrying capacity of upward direction of bus line 2.}
\begin{tabular}{
>{\columncolor[HTML]{FFFFFF}}l
>{\columncolor[HTML]{FFFFFF}}l
>{\columncolor[HTML]{FFFFFF}}l
>{\columncolor[HTML]{FFFFFF}}l
>{\columncolor[HTML]{FFFFFF}}l
>{\columncolor[HTML]{FFFFFF}}l
>{\columncolor[HTML]{FFFFFF}}l
>{\columncolor[HTML]{FFFFFF}}l
>{\columncolor[HTML]{FFFFFF}}l }
\hline
\multicolumn{2}{l}{\cellcolor[HTML]{FFFFFF}Sampling rate} & 0.5 & 0.7 & 0.9 & 1.1 & 1.3 & 1.5 & 1.7 \\ \hline
\cellcolor[HTML]{FFFFFF}                           & ND   & 78  & 78  & 78  & 78  & 78  & 78  & 78  \\
\cellcolor[HTML]{FFFFFF}                           & AWT  & 5.6 & 5.6 & 5.5 & 5.6 & 5.6 & 5.7 & 5.6 \\
\multirow{-3}{*}{\cellcolor[HTML]{FFFFFF}GA}       & NSP  & 0   & 0   & 0   & 0   & 0   & 0   & 0   \\ \hline
\cellcolor[HTML]{FFFFFF}                           & ND   & 66  & 66  & 66  & 66  & 66  & 66  & 66  \\
\cellcolor[HTML]{FFFFFF}                           & AWT  & 7.1 & 7.1 & 7.1 & 7.2 & 7.2 & 7.1 & 7.2 \\
\multirow{-3}{*}{\cellcolor[HTML]{FFFFFF}BTOA-MA}  & NSP  & 0   & 0   & 0   & 0   & 0   & 0   & 1   \\ \hline
\cellcolor[HTML]{FFFFFF}                           & ND   & 72  & 72  & 72  & 72  & 72  & 72  & 72  \\
\cellcolor[HTML]{FFFFFF}                           & AWT  & 6.2 & 6.1 & 6.2 & 6.2 & 6   & 6.2 & 6.3 \\
\multirow{-3}{*}{\cellcolor[HTML]{FFFFFF}Manual}   & NSP  & 0   & 0   & 0   & 0   & 0   & 0   & 0   \\ \hline
\cellcolor[HTML]{FFFFFF}                           & ND   & 48  & 52  & 55  & 59  & 63  & 65  & 70  \\
\cellcolor[HTML]{FFFFFF}                           & AWT  & 9.3 & 8.3 & 7.6 & 7.1 & 6.7 & 6.2 & 6   \\
\multirow{-3}{*}{\cellcolor[HTML]{FFFFFF}DRL-TO}   & NSP  & 0   & 0   & 0   & 0   & 0   & 0   & 0   \\ \hline
\end{tabular}
\end{table}

\begin{figure}[h]
\centering
\includegraphics[height=12cm,width=9cm]{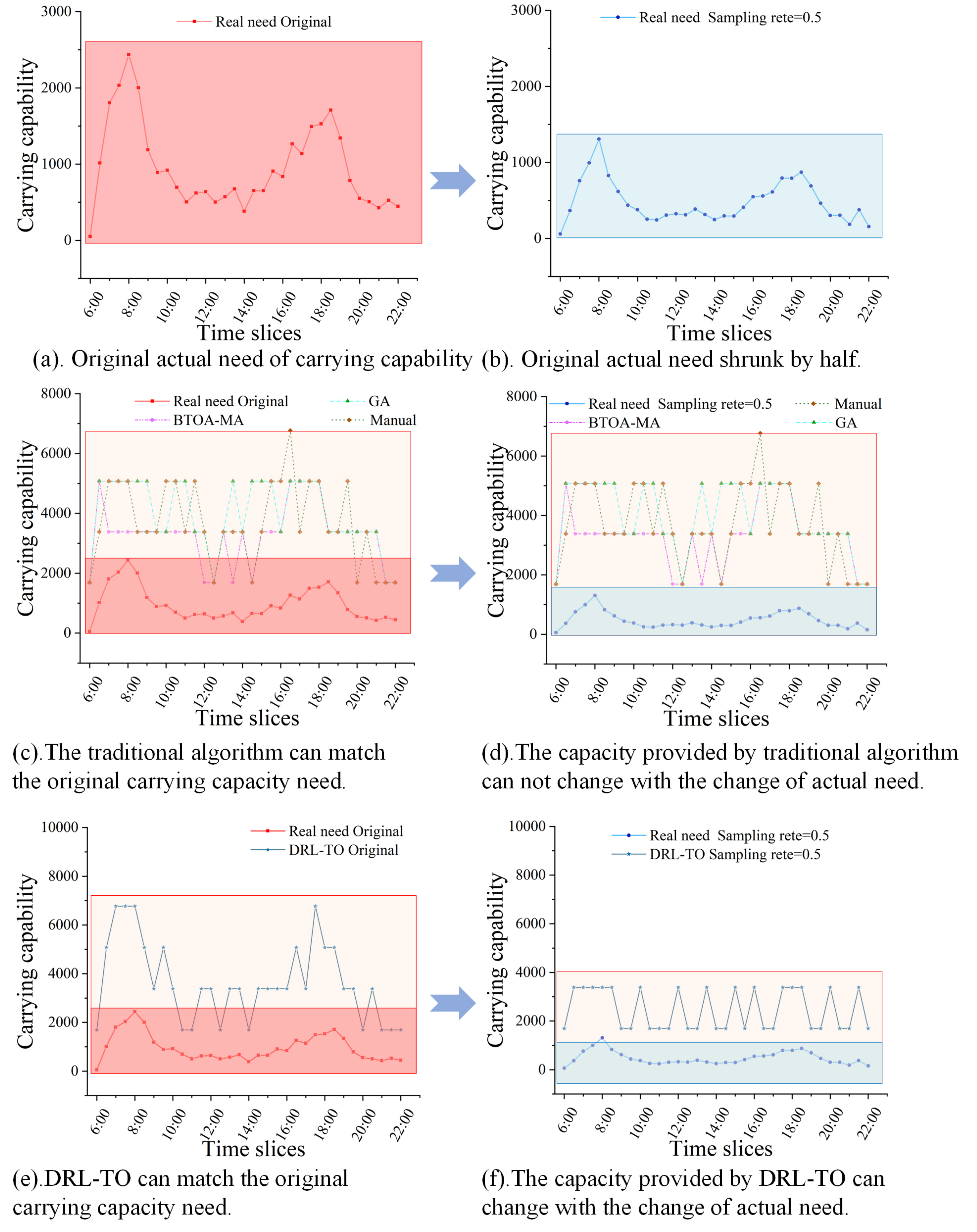}
\centering
\caption{Different time shifts on the peak of passenger flow for the upward direction of bus line 230 in time dimension.}
\end{figure}

Fig. 8 and Table \uppercase\expandafter{\romannumeral5} show the ability of DRL-TO, GA, BTOA-MA and Manual to handle both the increasing and decreasing levels of the passenger flow. As shown in Fig. 8 (a) and (b), we randomly sample the passenger swiping data with a sampling rate of 0.5. From the comparison of Fig. 8 (e) and Fig. 8 (f), it can be seen that when the carrying capacity need is reduced, DRL-TO can adjust the departure timetable in real-time. Table \uppercase\expandafter{\romannumeral5}  shows that the number of departures produced by DRL-TO in real time matches consistently with the change of sampling rate.  As the passenger flow sampling rate increases, DRL-TO will increase the departure frequency accordingly and the waiting time for passengers is shortened. However, other algorithms cannot precisely adjust the departing interval according to changes in passenger flow, which also can be seen from the comparison results in Fig. 8 (c) and (d). When the passenger need for carrying capacity drops, the carrying capacity of the timetable generated by other algorithms remain unchanged. That can also be verified in Table \uppercase\expandafter{\romannumeral5}, when the passenger flow changes, the times of departure in timetable generated by other algorithms and the average waiting time of passengers remain unchanged.

\subsubsection{Experimental results of DRL-TO parameter analysis}

\begin{table}[]
\caption{Experimental results of reward function parameter $\omega$ analysis.}
\begin{tabular}{lllllll}
\hline
\multirow{2}{*}{$\omega$} & \multicolumn{3}{l}{ND (Number of   Departures)} & \multicolumn{3}{l}{AWT (Average   Waiting Time)} \\ \cline{2-7}
                   & max            & min           & mode           & max           & min           & average          \\ \hline
0                  & 62             & 59            & 60             & 8             & 7.1           & 7.6              \\
1/11000            & 69             & 61            & 64             & 7.7           & 6.7           & 7                \\
1/9000             & 68             & 62            & 64             & 7.2           & 6.5           & 6.8              \\
1/7000             & 68             & 61            & 65             & 7.7           & 6.5           & 6.9              \\
1/5000             & 72             & 62            & 67             & 8             & 6.2           & 6.7              \\
1/4000             & 72             & 65            & 67             & 6.8           & 6.1           & 6.4              \\
1/3000             & 73             & 68            & 70             & 6.7           & 6             & 6.2              \\
1/2000             & 78             & 70            & 74             & 7.2           & 5.5           & 5.9              \\
1/1500             & 82             & 75            & 79             & 6.2           & 5.9           & 5.4              \\
1/1000             & 92             & 84            & 86             & 5.6           & 4.6           & 4.9              \\
1/500              & 107            & 98            & 102            & 4.9           & 4             & 4.1              \\ \hline
\end{tabular}
\end{table}

Regarding reward function is proposed in Eq. (2), there are two adjustable parameters $\omega$  and $\beta$. Among them,  $\beta$ is the parameter that controls the penalty due to stranding passengers.  $\omega$ controls the relative weight between waiting time and capacity matching. In other words, it controls the relative weighting between the cost of the bus company and the quality of service. It is necessary to investigate the influence of  $\omega$ on the interests of bus company and passengers. We adopt AWT to quantify the service quality, and ND to reflect the interests of the bus company.

\begin{figure}[htbp]
\centering
\includegraphics[height=4.7cm,width=6cm]{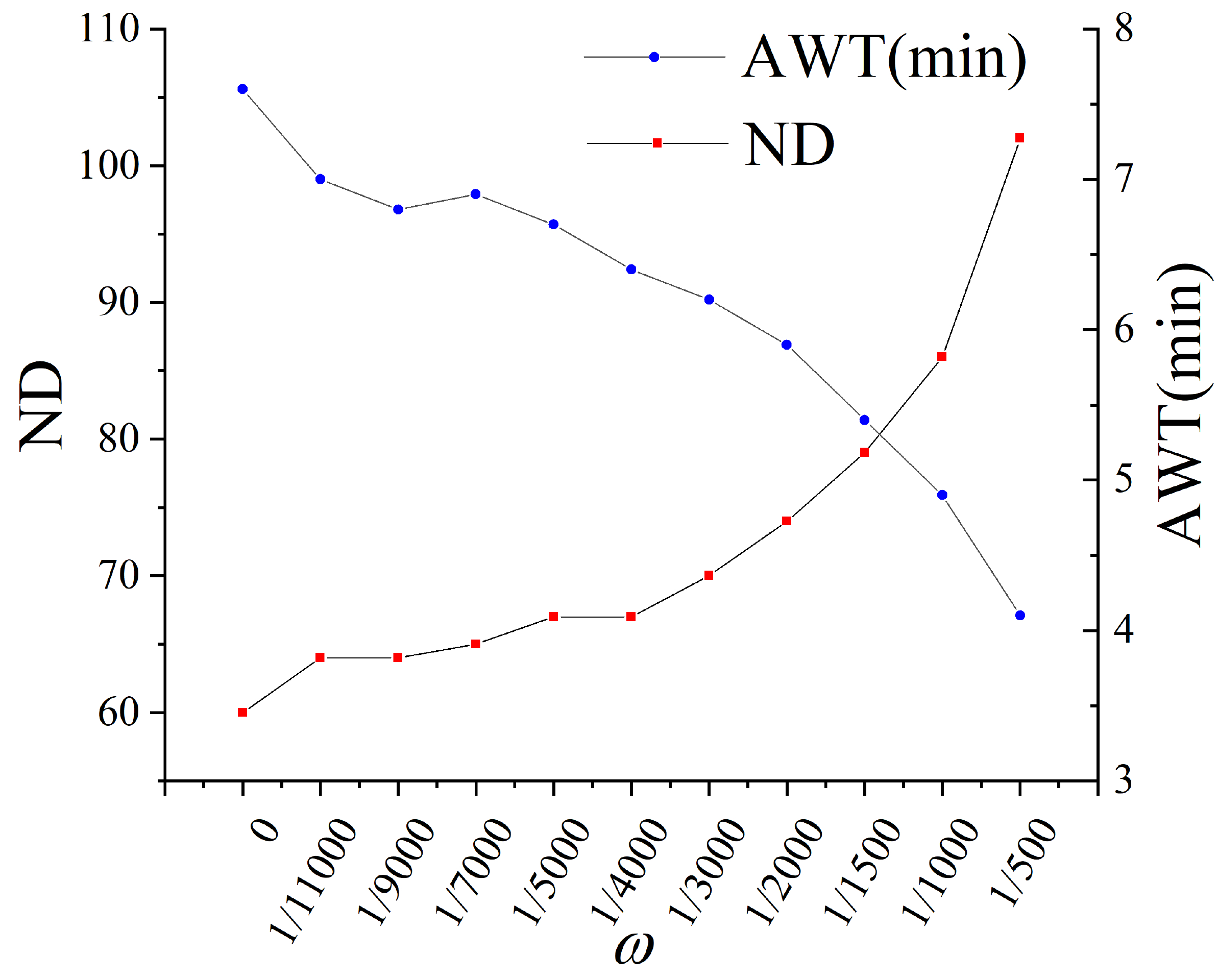}
\centering
\caption{The influence of parameter $\omega$ on the number of departures and waiting time.}
\end{figure}

We conducted experiments regarding the downward direction of line 2 to verify the influence of  $\omega$. We keep other parameters unchanged, adjust the values of  $\omega$, and observe the changes in AWT and ND. Table \uppercase\expandafter{\romannumeral6}  and Fig. 9 both confirm that that, upon increasing  $\omega$, the number of departures tends to increase, and the waiting time for passengers tends to decrease. When  $\omega$ is getting larger, more frequent bus departure can be witnessed in order to reduce AWT.

\begin{figure}[htbp]
\centering
\includegraphics[height=4.7cm,width=6cm]{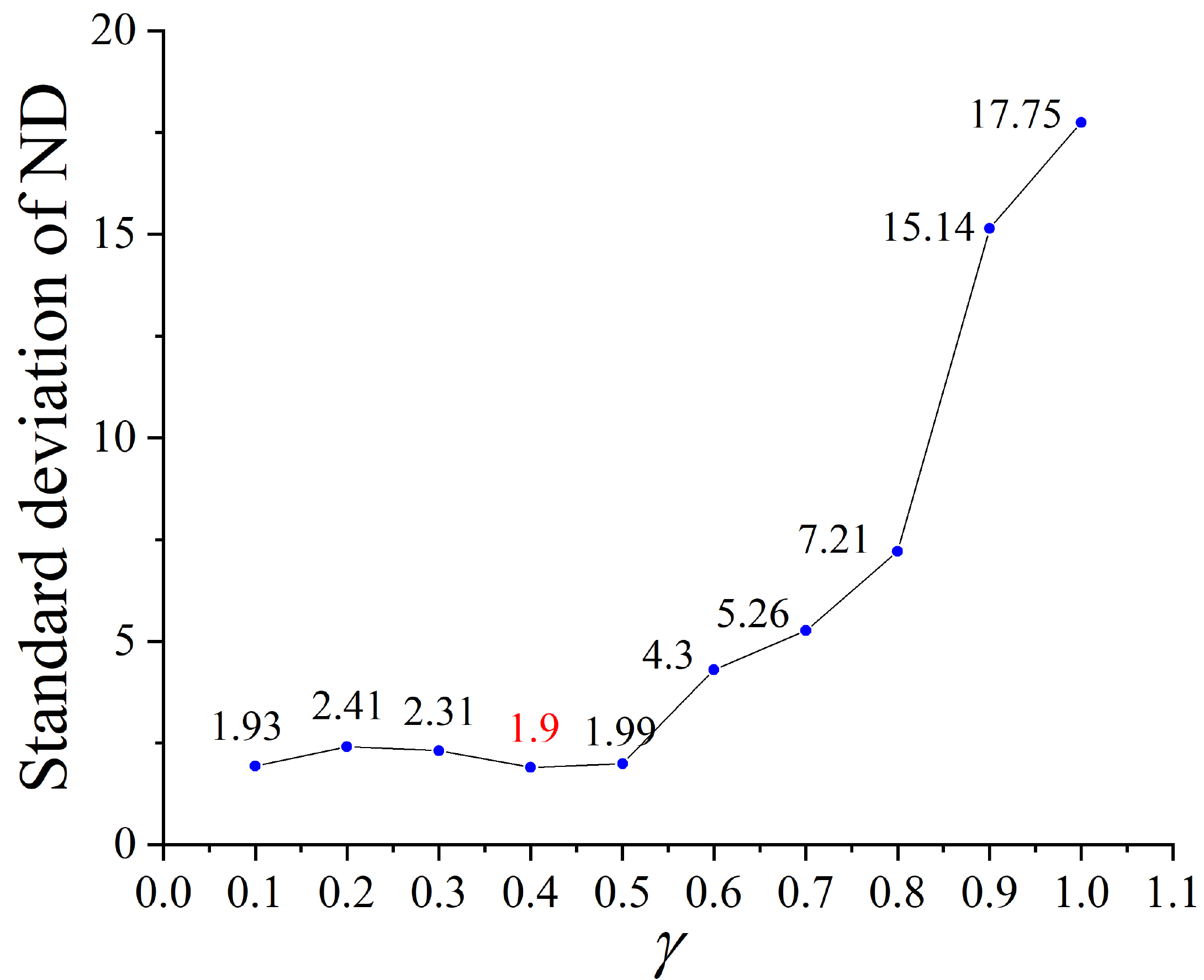}
\centering
\caption{The influence of parameter $\gamma$  on stability of convergence results.}
\end{figure}

The degree of fluctuation in the number of departures in bus timetable reflects the stability of DRL-TO. The more stable DRL-TO, the higher the reliability of the timetable provided by it in practical applications. We study the influence of the discount reward factor $\gamma$  on the stability of the departure number achieved by DRL-TO on the downward direction of line 2. We report the standard deviation of the number of departures observed across 25 independent episodes. As shown in Fig. 10, when  $\gamma= 0.4$, the standard deviation of departure number is the smallest, so we set   $\gamma= 0.4$ in this paper.

\subsubsection{Experimental results of ablation study}

To demonstrate the usefulness of our state space design in subsection \uppercase\expandafter{\romannumeral4} \emph{D} and the definition of the reward function in subsection \uppercase\expandafter{\romannumeral4} \emph{C}, we conduct a series of experiments regarding on the upward direction data of line 239, and compare our design with two other state and reward function design schemes.

Scheme one:
We concatenate state features with respect to each station to form the complete state description. Specifically: ${s_e} = [s_e^{_{m - 3}},s_e^{_{m - 2}},s_e^{_{m - 1}},s_e^{_m}]$, where  $s_e^m = [s_e^{m,1},s_e^{m,2},s_e^{m,3},s_e^{m,4}]$, $s_e^{m,1},s_e^{m,2},s_e^{m,3}$  and $s_e^{m,4}$ are all vectors of length $K-1$,  $s_e^{m,1}$ represents the number of passengers waiting at each station at the  $m-th$ minute,  $s_e^{m,2}$ indicates the number of passengers arriving at each station in the next 15 minutes,  $s_e^{m,3}$ is the location of each bus. When a bus is running between the $k-th$ station and the $(k+1)-th$  station, the dimension of  $s_e^{m,3}$  is 1. The $k-th$  dimension of $s_e^{m,4}$  represents the distance of the bus from the $k-th$  station when the bus is located in between the $k-th$  station and the $(k+1)-th$  station. If there are multiple buses, the closest one to the $(k+1)-th$  station is utilized. This is because it has the biggest impact on the next station.  $s_e^{m,4}$ is expressed by a fraction, which is the ratio of the distance with respect to the bus leaving the $k-th$   station to the distance between the  $k-th$  station and the $(k+1)-th$  station.

Taking into account the interests of both the bus company and the passengers, the ideal state is: the bus is fully loaded, and the waiting passengers at each station is 0. So we designed the following reward function:

\begin{equation}\label{equ15}
r =  - ({C_{ml}} + \sum\nolimits_{k = 1}^{k = K} {{d_{mk}}} )
\end{equation}
where ${C_{ml}}$  represents the total number of remaining seats on the bus running on the line at the $m-th$  minute, and  ${d_{mk}}$ is the number of passengers waiting for the bus at the $k-th$  station at the $m-th$  minute.

Scheme two:
In scheme 1, only the state of the first station is changed by the departing action directly, while the states of other stations are changed by all the on-road buses and the passenger flow. Therefore, we design a new state description for scheme two. First, we segment the passenger flow into consecutive hours, and use $t_h$  to indicate the hourly time mark. Secondly, in order to narrow the gap between the passenger need and the service provided by the bus company, we select the total number of passengers carried at each station without limiting the maximum number of passengers onboard. In addition, the time interval between the current time and the previous departure is considered. Therefore, we define  ${s_{{e'}}} = [{t_h},max{\kern 1pt} p_m^k,t_m^l]$,  where $t_h$   represents the hourly period that the $m-th$  minute belongs to, $t_m^l$  represents the interval between the time when the previous vehicle departure from the starting station and the $m-th$  minute.

\begin{equation}\label{equ16}
p_m^k = \sum\nolimits_{k = 1}^{k = k} {(w_m^k}  - h_m^k)
\end{equation}
where $w_m^k$  is the number of passengers waiting at the $k-th$  station at the $m-th$  minute, $h_m^k$   represents the number of passengers who got off at the $k-th$  station, and $p_m^k$   indicates the number of passengers carried when the bus departed in the $m-th$  minute leaves the $k-th$  station.
To match the state space, we define the reward function as follows:

\begin{equation}\label{equ17}
r =  - [{e_m} - \sum\nolimits_{k = 1}^{k = K} ( w_m^k - h_m^k)] - \sum\nolimits_{k = 1}^{k = K} {w_m^k}
\end{equation}
where $e_m$  is the carrying capacity provided by the bus sent out in the $m-th$  minute, $\sum\nolimits_{k = 1}^{k = K} ( w_m^k - h_m^k)$  means the capacity consumed from the bus sent out in the $m-th$  minute, and $\sum\nolimits_{k = 1}^{k = K} {w_m^k}$  is the number of passengers waiting for the bus.

\begin{figure}[htbp]
\centering
\includegraphics[height=6.7cm,width=8cm]{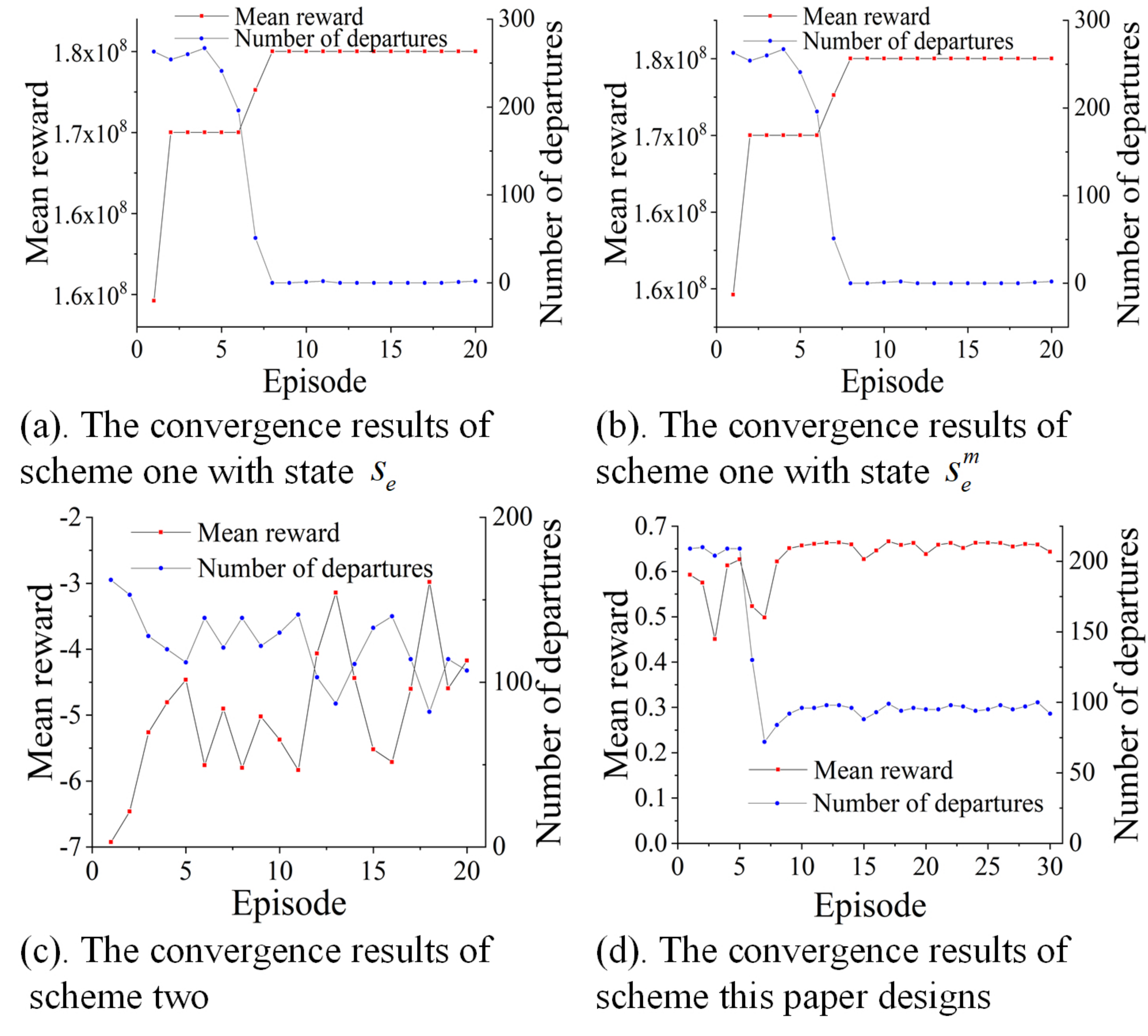}
\centering
\caption{Convergence comparison of each scheme.}
\end{figure}

Fig. 11 (a) and (b) show the average reward per episode of DQN, based on scheme 1. For the same reward function, we tried to use ${s_e}$ and ${s_e^m} $ as the environment state. Experiments show that the number of departures achieved by scheme one consistently approaches to 0. As far as the bus system is concerned, the only action is to departure or not. The action has a large delay in changing other states. In addition, new passengers may be arrive at every station from time to time, which prevents the action from imposing big changes to the state. It may be an important reason that reinforcement learning algorithms cannot make correct decisions.

As shown in Fig. 11(c), DQN can converge under scheme two, but the result is not stable. The reason for this situation may be that there is no restriction on the maximum number of passengers on the bus, so the state feature $\max p_m^k$  is unstable. and only the scheme proposed in this paper can converge to a desirable solution, which can be verified by Fig. 11(d).

\begin{table*}[]
\caption{State Features Necessity Analysis Results.}
\begin{tabular}{llllllll}
\hline
                                         & Feature  & $s_m$                                    & ${s_m}{\rm{ - }}{x_{1m}}{\rm{ - }}{x_{2m}}$        & ${s_m}{\rm{ - }}{x_{3m}}$       &${s_m}{\rm{ - }}{x_{4m}}$       &${s_m}{\rm{ - }}{x_{5m}}$       & ${s_m}{\rm{ - }}{ds_{m}}$        \\ \hline
                                         & max      & 0.749                                & 0.748    & 0.749    & 0.742    & 0.748    & 0.749    \\
                                         & min      & 0.741                                & 0.742    & 0.74     & 0.729    & 0.734    & 0.741    \\
\multirow{-3}{*}{Mean reward}            & variance & 3.74E-06                             & 3.51E-06 & 6.03E-06 & 1.41E-05 & 1.33E-05 & 4.53E-06 \\ \hline
                                         & max      & 110                                  & 111      & 111      & 112      & 110      & 111      \\
                                         & min      & 106                                  & 107      & 107      & 107      & 105      & 106      \\
                                         & mode     & 109                                  & 110      & 109      & 110      & 108      & 108      \\
\multirow{-4}{*}{Number of   departures} & variance & {\color[HTML]{FE0000} \textbf{0.94}} & 0.99     & 1.35     & 2.87     & 1.5      & 1.41     \\ \hline
\end{tabular}
\end{table*}

In order to analyze the necessity of each state selection, we add and remove some state features. As shown in Table \uppercase\expandafter{\romannumeral7}, $s_m$  represents the state space designed in \uppercase\expandafter{\romannumeral4} \emph{D}, and ${s_m}{\rm{ - }}{x_{1m}}{\rm{ - }}{x_{2m}}$  represents the deletion of the temporal feature from the state space defined in Eq. (4), ${s_m}{\rm{ - }}{x_{3m}}$, ${s_m}{\rm{ - }}{x_{4m}}$, ${s_m}{\rm{ - }}{x_{5m}}$ and ${s_m}{\rm{ - }}{ds_{m}}$   respectively indicate the deletion of the maximum full load rate, the waiting time for passengers, the utilization rate of the capacity, and the addition of the number of stranding passengers.

We perform experiments to determine the maximum, minimum, mode, and variance of the number of departures and the average reward of the 20 decision-making sequences on upward direction of line 239, after the learning process of DRL-TO converges. As shown in Table \uppercase\expandafter{\romannumeral6}, the new state description with less or more features are compared with the state space defined in Eq. (4). The bus timetable generated based on state space in Eq. (4) can produce the smallest difference between the maximum value and the minimum numbers of bus departures as well as the smallest variance. Therefore, the state space designed in Section 4.4 is the most stable, which can also be verified by the small variance of the average reward.

\section{Conclusion}

In order to meet the demand of suddenly changed passenger flows, this paper proposes a dynamic generation method of bus timetable based on deep reinforcement learning. We regard the bus timetable optimization problem as a sequential decision problem, and uses reinforcement learning to determine whether to depart at the current time, thus to determine the departure interval in real time. We select the load factor, carrying capacity utilization rate, the number of stranding passengers as the state of reinforcement learning, and whether departure as the action. Taking into account the interests of both the bus company and passengers, a reward function is designed, which including the indicators of full load rate, empty load rate, waiting time, and the number of stranding passengers. We improve the carrying capacity calculation method in our previous work by ameliorating the matching degree of capacity at each station.

Experiments demonstrate that compared with the timetable generated by state-of-the-art algorit hms BTOA-MA, GA and the manual, DRL-TO can control departure interval by determining whether to departure every minute. The departure timetable can be dynamically generated based on the real-time passenger flow. As a result, the bus timetable generated by DRL-TO saves 8\% of the bus number and reduces 17\% of passengers wait time on average.

\appendices
\section{Proof of the First Zonklar Equation}
Appendix one text goes here.


The authors would like to thank...

\ifCLASSOPTIONcaptionsoff
  \newpage
\fi



%


%

\begin{IEEEbiography}[{\includegraphics[width=1in,height=1.25in,clip,keepaspectratio]{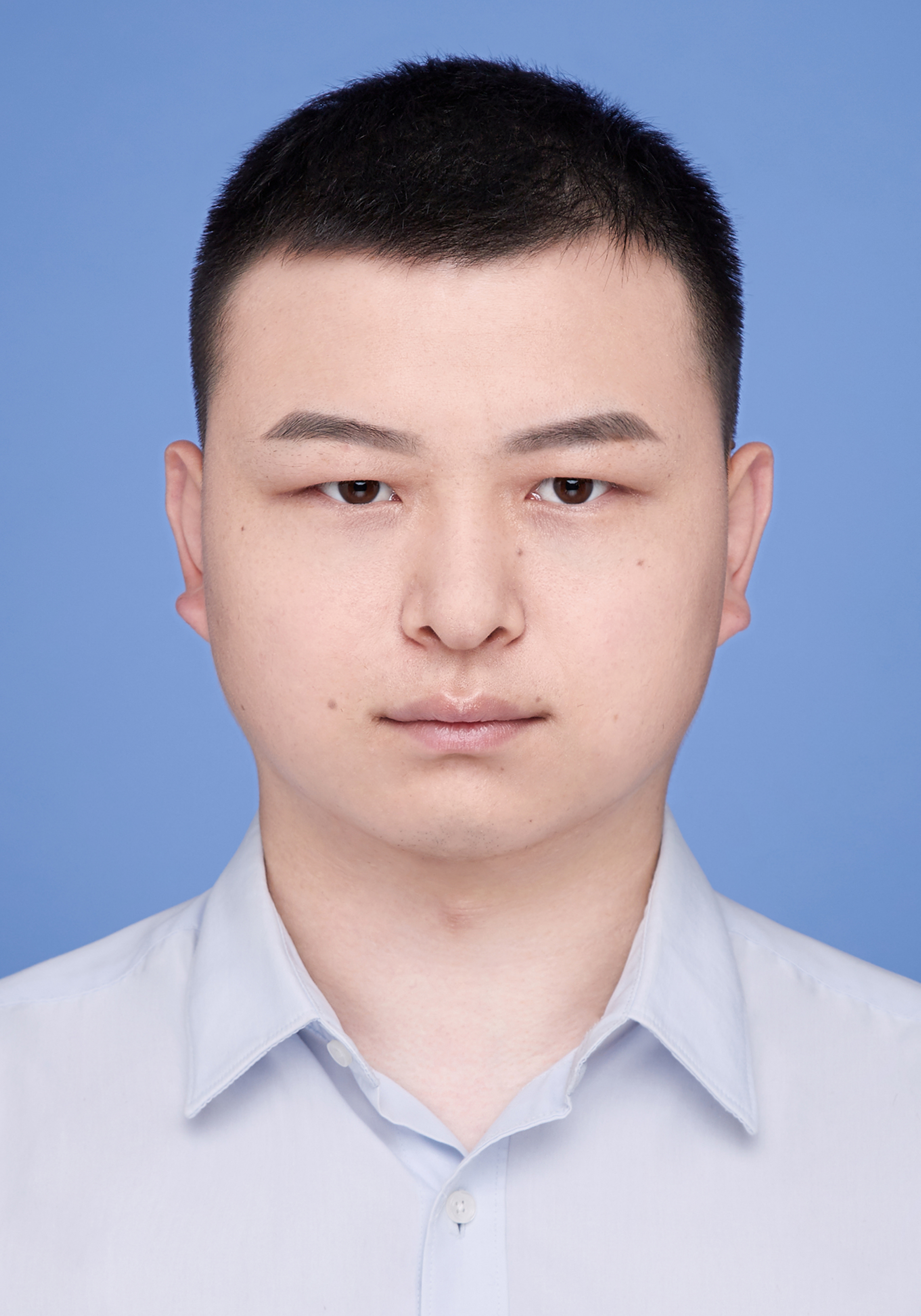}}]{Guanqun Ai}
is currently pursuing the Ph.D. degree in Computer Science and Technology from Beijing University of Posts and Telecommunications, Beijing, China. He received the masters degree in School of Automation from Beijing University of Posts and Telecommunications in 2020. His research interests include bus timetable optimization, reinforcement learning and data mining.
\end{IEEEbiography}

\begin{IEEEbiography}[{\includegraphics[width=1in,height=1.25in,clip,keepaspectratio]{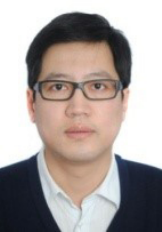}}]{Xingquan Zuo}
(SM$\sim$14) is currently a Professor in School of Computer Science, Beijing University of Posts and Telecommunications. He received the Ph.D. degree in Control Theory and Control Engineering from Harbin Institute of Technology, Harbin, China, in 2004. From 2004 to 2006, he was a Postdoctoral Research Fellow in Automation Department of Tsinghua University. From 2012 to 2013, he was a Visiting Scholar in Industrial and System Engineering Department, Auburn University, AL, USA. His research interests are in optimization and scheduling, evolutionary computation, data mining and intelligent transportation systems. He has published over 100 research papers in journals and conferences, two books and several book chapters.

He is a Senior Member of IEEE, Member of ACM, Committee Member of Intelligent Service Society of Chinese Association for Artificial Intelligence, Committee Member of Transportation Model and Simulation Society and Intelligent Simulation Optimization and Scheduling Society of Chinese Association for System Simulation, and Senior Member of Chinese Association for Artificial Intelligence. He served in Program Committee of more than 10 international conferences.
\end{IEEEbiography}

\begin{IEEEbiography}[{\includegraphics[width=1in,height=1.25in,clip,keepaspectratio]{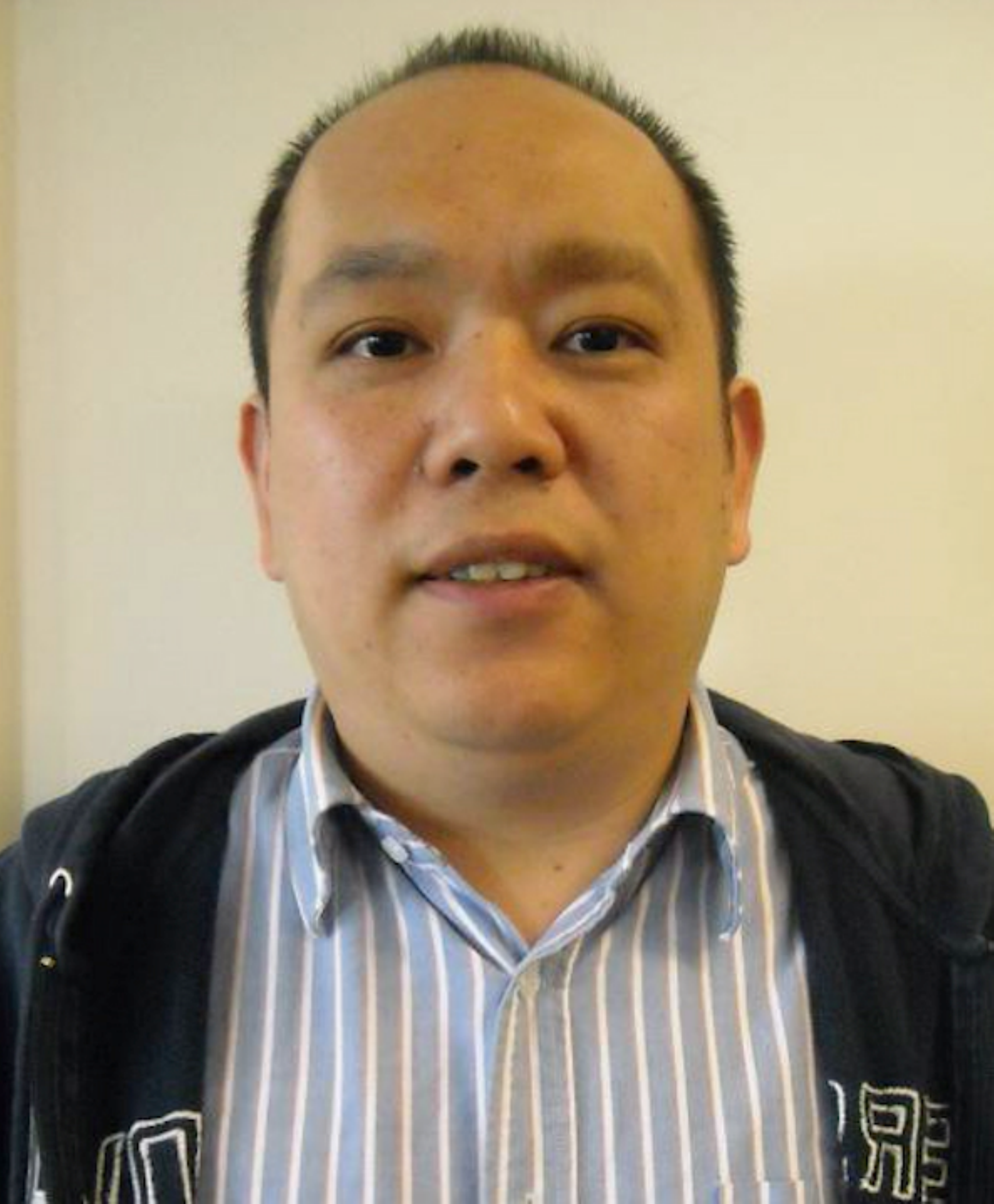}}]{Gang Chen}
 obtained his B.Eng degree from Beijing Institute of Technology in China and PhD degree from Nanyang Technological University (NTU) in Singapore respectively. He is currently a senior lecturer in the School of Engineering and Computer Science at Victoria University of Wellington. His research interests include evolutionary computation, reinforcement learning, multi-agent systems and cloud and service computing. He has more than 120 publications, including leading journals and conferences in machine learning, evolutionary computation, and distributed computing areas, such as IEEE TPDS, IEEE TEVC, JAAMAS, ACM TAAS, IEEE ICWS, IEEE SCC. He is serving as the PC member of many prestigious conferences including ICLR, ICML, NeurIPS, IJCAI, and AAAI, and co-chair for Australian AI 2018 and CEC 2019.
\end{IEEEbiography}

\begin{IEEEbiography}[{\includegraphics[width=1in,height=1.25in,clip,keepaspectratio]{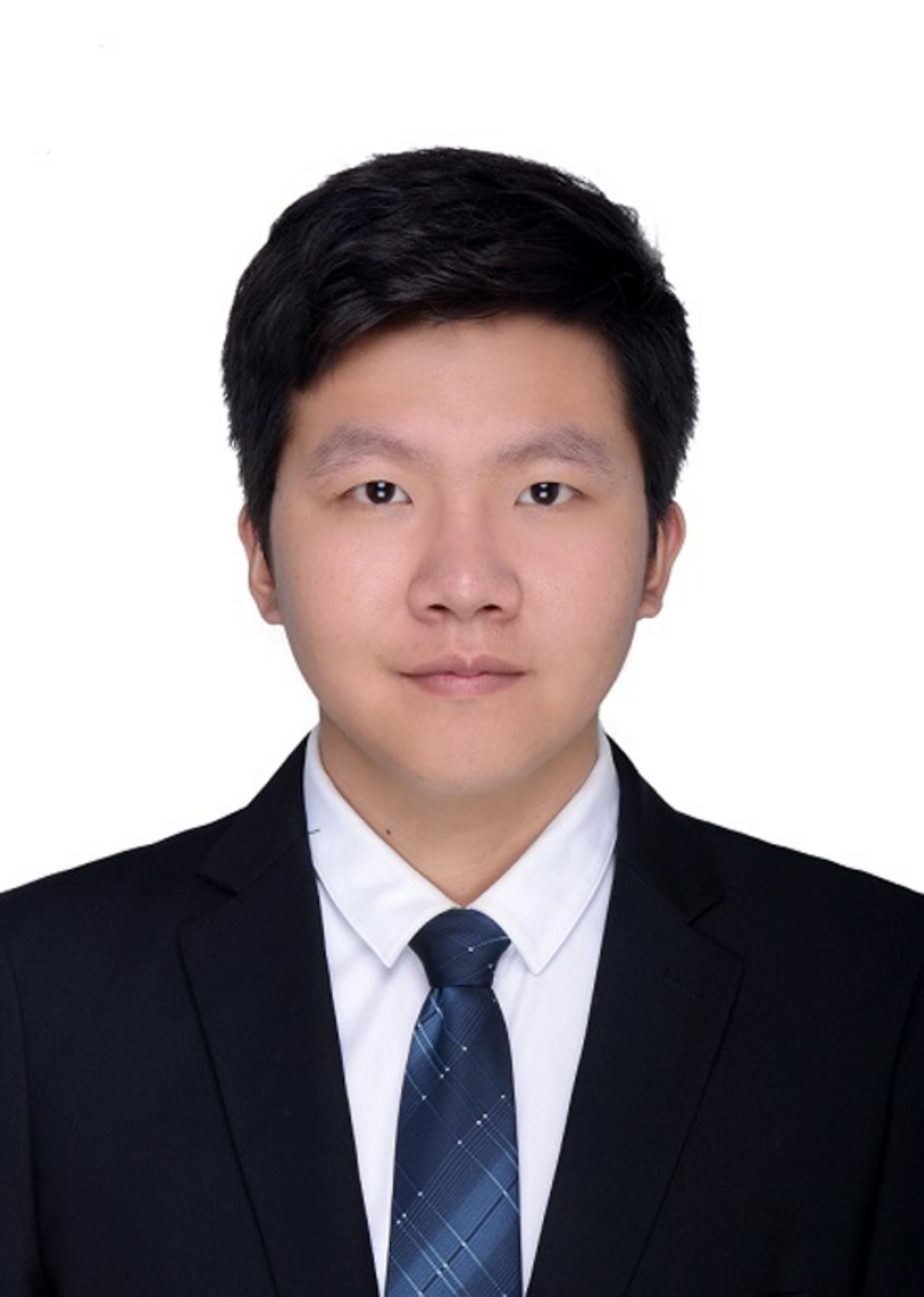}}]{Binglin Wu}
 is currently pursuing the Ph.D. degree in Computer Science and Technology from Beijing University of Posts and Telecommunications, Beijing, China. He received the masters degree in School of Information Engineering from Zhengzhou University, Zhengzhou, China, in 2018. His research interests include bus network design, evolutionary computation and reinforcement learning with applications.
\end{IEEEbiography}




\end{document}